\pdfoutput=1
\documentclass[11pt]{article}

\usepackage[]{ACL2023}

\usepackage{times}
\usepackage{latexsym}

\usepackage{amsmath}
\usepackage{amssymb}
\usepackage{makecell}
\usepackage{caption}

\usepackage[T1]{fontenc}
\usepackage[utf8]{inputenc}

\title{Open-SQL Framework: Enhancing Text-to-SQL on Open-source Large Language Models}
\author{304343992 }
\date{February 2024}

\usepackage{subfigure}
\usepackage{xcolor}
\usepackage{xspace}
\usepackage{booktabs}
\usepackage{url}
\usepackage{multirow}
\usepackage{listings}
\usepackage{makecell}
\usepackage{tabularx}
\usepackage{utfsym}
\usepackage[linesnumbered, boxed]{algorithm2e}
\usepackage{amsmath, amsthm}
\usepackage{pifont}
\usepackage{acl}
\usepackage{mylstlinebgrd}
\usepackage{graphicx}
\usepackage{hyperref}
\usepackage{comment}

\newcommand{\lmc}{Llama 2\xspace}
\newcommand{\clm}{Code Llama\xspace}

\newcommand{\bsprompt}{Basic Prompt\xspace}
\newcommand{\textprompt}{Text Representation Prompt\xspace}
\newcommand{\sqlprompt}{Code Representation Prompt\xspace}
\newcommand{\openaiprompt}{OpenAI Demostration Prompt\xspace}
\newcommand{\alpacaprompt}{Alpaca SFT Prompt\xspace}
\newcommand{\openprompt}{Open Prompt\xspace}

\newcommand{\openschema}{Open DB Schema\xspace}
\newcommand{\openexample}{Open Example Curation\xspace}

\newcommand{\ours}{Open-SQL\xspace}

\definecolor{eclipseBlue}{RGB}{42,0.0,255}
\definecolor{eclipseGreen}{RGB}{63,180,95}
\definecolor{eclipsePurple}{RGB}{175,0,25}
\definecolor{codewhite}{rgb}{0.70,0.70,0.70}
\definecolor{codegray}{rgb}{0.5,0.5,0.5}
\definecolor{codepurple}{rgb}{0.58,0,0.82}
\definecolor{backcolour}{rgb}{0.95,0.95,0.92}

\definecolor{instructioncolor}{rgb}{0.6275, 0.7686, 0.6157}
\definecolor{taskcolor}{rgb}{0.8824, 0.9255, 0.7843}
\definecolor{prefixcolor}{rgb}{0.9922, 1, 0.6824}

\lstdefinelanguage{Prompt}{
	backgroundcolor=\color{backcolour},   
	keywordstyle=\color{magenta},
	numberstyle=\tiny\color{codegray},
	basicstyle=\ttfamily\footnotesize,
	breakatwhitespace=false,         
	breaklines=true,   
    breakindent=-3.5pt,
	captionpos=b,                    
	keepspaces=true,                 
	numbers=left,                    
	numbersep=5pt,                  
	showspaces=false,                
	showstringspaces=false,
	showtabs=false,                  
	tabsize=4,
	escapeinside={`}{`},
	morecomment = [s][\color{eclipseGreen}\bfseries]{How}{?},
        morecomment = [l][\color{eclipseBlue}\bfseries]{SELECT},
        morecomment = [l][\color{eclipsePurple}\bfseries]{\$\{DATABASE_SCHEMA\}},
        morecomment = [l][\color{eclipsePurple}\bfseries]{\$\{DATABASE_DEFN\}},
        morecomment = [l][\color{eclipsePurple}\bfseries]{\$\{EXTERNAL_KNOWLEDGE\}},
        morecomment = [l][\color{eclipsePurple}\bfseries]{\$\{TARGET_QUESTION\}},
        morecomment = [s][\color{eclipsePurple}\bfseries]{CREATE}{;},
        morecomment = [l][\color{eclipsePurple}\bfseries]{Table},
        morecomment = [l][\color{eclipsePurple}\bfseries]{continents},
        morecomment = [l][\color{eclipsePurple}\bfseries]{countries},
    postbreak={
       \mbox{
           \lst@linebreakbgrd
           \rotatebox[y=0.7ex]{180}{\color{black}$\Lsh\,$}
       }
    },
}

\lstdefinelanguage{Schema}{
	backgroundcolor=\color{backcolour},   
	keywordstyle=\color{magenta},
	numberstyle=\tiny\color{codegray},
	basicstyle=\ttfamily\footnotesize,
	breakatwhitespace=false,         
	breaklines=true,   
    breakindent=-3.5pt,
	captionpos=b,                    
	keepspaces=true,                 
	numbers=left,                    
	numbersep=5pt,                  
	showspaces=false,                
	showstringspaces=false,
	showtabs=false,                  
	tabsize=4,
	escapeinside={`}{`},
	morecomment = [l][\color{eclipseBlue}\bfseries]{(},
        morecomment = [l][\color{eclipseBlue}\bfseries]{)},
    postbreak={
       \mbox{
           \lst@linebreakbgrd
           \rotatebox[y=0.7ex]{180}{\color{black}$\Lsh\,$}
       }
    },
}

\lstdefinelanguage{Question}{
	backgroundcolor=\color{backcolour},   
	sensitive = true,
	morecomment = [s]{Gold}{SQL},
	commentstyle ={\color{red}\bfseries\underbar},
	morestring = [b]",
	morestring = [b]',
	stringstyle = \color{eclipseGreen},
	basicstyle=\ttfamily\footnotesize,
	breaklines=true,
	alsoletter=!?-,
	emph={Question},
	emphstyle={\color{red}\bfseries\underbar},
	captionpos=b,
	escapeinside={[}{]},
	keepspaces=true,              
	showspaces=false,                
	showstringspaces=false,
	showtabs=false,                  
	tabsize=4,
	columns=flexible
}

\begin{document}

\author{
Xiaojun Chen$^*$ \qquad Tianle Wang$^*$ \qquad Tianhao Qiu$^*$ \\
\textbf{Jianbin Qin$^*$} \qquad \textbf{Min Yang$^\dag$} \\
\\
$^*$ Shenzhen University, Shenzhen, China \\
$^\dag$ Shenzhen Institute of Advanced Technology (SIAT), Chinese Academy of Sciences (CAS), Shenzhen, China \\
xjchen@szu.edu.cn, 2210273001@email.szu.edu.cn, 2310275033@email.szu.edu.cn, \\ qinjianbin@szu.edu.cn, min.yang@siat.ac.cn
}

\maketitle

\begin{abstract}
Despite the success of large language models (LLMs) in Text-to-SQL tasks, open-source LLMs encounter challenges in contextual understanding and response coherence. To tackle these issues, we present \ours, a systematic methodology tailored for Text-to-SQL with open-source LLMs. Our contributions include a comprehensive evaluation of open-source LLMs in Text-to-SQL tasks, the \openprompt strategy for effective question representation, and novel strategies for supervised fine-tuning. We explore the benefits of Chain-of-Thought in step-by-step inference and propose the \openexample method for enhanced few-shot learning. Additionally, we introduce token-efficient techniques, such as \textbf{Variable-length Open DB Schema}, \textbf{Target Column Truncation}, and \textbf{Example Column Truncation}, addressing challenges in large-scale databases. Our findings emphasize the need for further investigation into the impact of supervised fine-tuning on contextual learning capabilities. Remarkably, our method significantly improved Llama2-7B from 2.54\% to 41.04\% and Code Llama-7B from 14.54\% to 48.24\% on the BIRD-Dev dataset. Notably, the performance of Code Llama-7B surpassed GPT-4 (46.35\%)~\footnote{https://bird-bench.github.io/} on the BIRD-Dev dataset.
\end{abstract}

\section{Introduction}
\label{sec:intro}

% background

SQL, efficient for professionals, may pose a barrier for non-technical users to access relational data. Text-to-SQL parsing, translating natural language questions into machine-executable SQL queries, has gained attention in both industry and academia~\cite{deng2022recent}. This approach empowers non-expert users to query tables and plays a pivotal role in applications such as customer service, question answering, and robotic navigation.

%related work
Most previous works~\cite{rat-sql, li2023graphix} focus on extracting the question-to-SQL patterns and generalizing them by training an encoder-decoder model with Text-to-SQL corpus. In recent years, Large Language Models (LLMs) have emerged as a new paradigm for Text-to-SQL~\cite{DBLP:journals/corr/abs-2303-13547}. Different from prior studies, the core problem in LLM-based Text-to-SQL solution is how to prompt LLM to generate correct SQL queries, namely prompt engineering. 
Such prompt engineering involves question representations \cite{pourreza2023dinsql}, example selection \cite{nan2023enhancing}, and example organization \cite{guo2023framework}.

% Open-source LLM is an alternative. 
% Challenge in Open-source LLM for Text-to-SQL. 

Recently, open-source LLMs have gained traction in programming and text tasks due to transparency, accessibility, affordability, privacy benefits, and community-driven development. Despite these advantages, they often lag behind OpenAI LLMs in context understanding and coherent response generation. A key challenge for open-source LLMs is improving performance in Text-to-SQL, achievable through supervised fine-tuning and in-context learning.

% summary
In this paper, we present a systematic methodology, labeled as \ours, crafted for Text-to-SQL tasks with open-source LLMs like Llama and Code Llama. This methodology provides effective and efficient techniques for both supervised fine-tuning and in-context learning. Our main contributions and results are summarized as follows: 
\begin{itemize}
    \item \textbf{Systematic Evaluation:} A comprehensive assessment of open-source LLMs on Text-to-SQL tasks using the BIRD dataset, revealing significant performance deficiencies in understanding the provided database schema.

    \item \textbf{Effective Strategies:} We introduced effective strategies, including the \textbf{\openprompt} question representation, two novel Chain-of-Thought (COT) methods for step-by-step inference improvement—the first exploration of COT in Text-to-SQL tasks—and \openexample for few-shot learning, enhancing performance across both GPT and open-source LLMs.
    
    \item \textbf{Token Efficiency for Large-scale Databases:} We introduce techniques to overcome memory limitations during fine-tuning and inference with few examples on large-scale databases by reducing token length. The proposed strategies, including \textbf{Variable-length Open DB Schema}, \textbf{Target Column Truncation}, and \textbf{Example Column Truncation}, verified through experiments, play significant roles in ensuring the success of supervised fine-tuning and few-shot learning for open-source LLMs in Text-to-SQL tasks.

    \item \textbf{Benefits of Supervised Fine-Tuning:} Experimental evidence demonstrates significant performance enhancements in open-source LLMs for Text-to-SQL tasks achieved through question representation strategies, supervised fine-tuning, Chain-of-Thought (COT), and few-shot learning. Significantly, our method boosted Llama2-7B from 2.54\% to 41.04\% and Code Llama-7B from 14.54\% to 48.24\% on the BIRD-Dev dataset. Notably, Code Llama-7B after fine-tuning surpassed GPT-4 (46.35\%), maintaining its leading position among open-source LLMs.

    \item \textbf{Open-Source Available:} We have open-sourced our code and models to foster further research, adoption, and innovation in the Text-to-SOL domain within the community~\footnote{\url{https://github.com/Bruce-XJChen/Open-SQL}}.
\end{itemize}

\section{Related Work}
\label{sec:related}

Recent studies have approached Text-to-SQL as a sequence-to-sequence task, employing machine learning models trained with encoder-decoder architectures~\cite{DBLP:conf/ijcai/CaiXZYLL18, DBLP:conf/coling/PopescuMVYKS22, qi2022rasat}. Techniques such as attention mechanisms, graph representation, syntax parsing, and deep learning have been extensively utilized to enhance Text-to-SQL methods~\cite{DBLP:conf/ijcnn/LiuSZWLK23, DBLP:conf/emnlp/XuWWFS18, li2023graphix, DBLP:conf/acl/ZhengWDWL22, rat-sql, qi2022rasat, DBLP:conf/acl/HuiGWQLLSL22, DBLP:conf/acl/GuoZGXLLZ19, scholak2021picard, li2023resdsql, wang2022proton}. The advent of pretraining language models, extended to Tabular Language Models (TaLMs) that directly encode tables and texts, has proven beneficial~\cite{yin2020tabert}. However, creating a versatile pretraining model specifically tailored for Text-to-SQL tasks remains a complex challenge with associated costs.

Recent advancements in natural language processing and artificial intelligence have witnessed the emergence of Large Language Models (LLMs) such as GPT from OpenAI \cite{chatgpt,gpt4} and Llama from Meta \cite{llama,llama2}. LLMs possess the potential to bridge the gap between natural language queries and structured SQL queries. Two prominent techniques for incorporating LLMs into the Text-to-SQL task include \textbf{supervised fine-tuning}, where additional Text-to-SQL instances refine LLMs for enhanced performance in specific tasks~\cite{sun2023sql}, and \textbf{in-context learning}~\cite{icl22}, also known as prompt engineering~\cite{nan2023enhancing, DBLP:journals/corr/abs-2303-13547} or few-shot learning. In the context of Text-to-SQL, various prompt construction methods have been proposed, involving question representations \cite{chang2023prompt,dong2023c3,openaiprompt,pourreza2023dinsql}, example selection \cite{guo2023framework,liu2022in-context,nan2023enhancing}, and example organization \cite{guo2023framework}.

In conclusion, open-source LLMs in Text-to-SQL tasks may provide cost savings and privacy benefits but exhibit lower proficiency in context comprehension and coherent response generation compared to proprietary counterparts, such as those from OpenAI, posing a notable challenge.
\section{Zero-shot Error Analysis}
\label{sec:error_analysis}

BIRD~\cite{DBLP:journals/corr/abs-2305-03111}, tailored for Text-to-SQL tasks in real-world applications, emphasizes data integrity, integrating natural language queries with database information, and improving SQL efficiency in large databases. It includes 9,427 training instances and 1,534 development instances across 95 databases in 37 domains, with each instance comprising a natural language question and its corresponding SQL query. For evaluation in this study, the development split, BIRD-dev, is utilized due to the unavailability of the test split.

\begin{figure}[t]
    \centering
    \subfigure[Result of Llama 2.]{
    \includegraphics[width=0.46\textwidth]{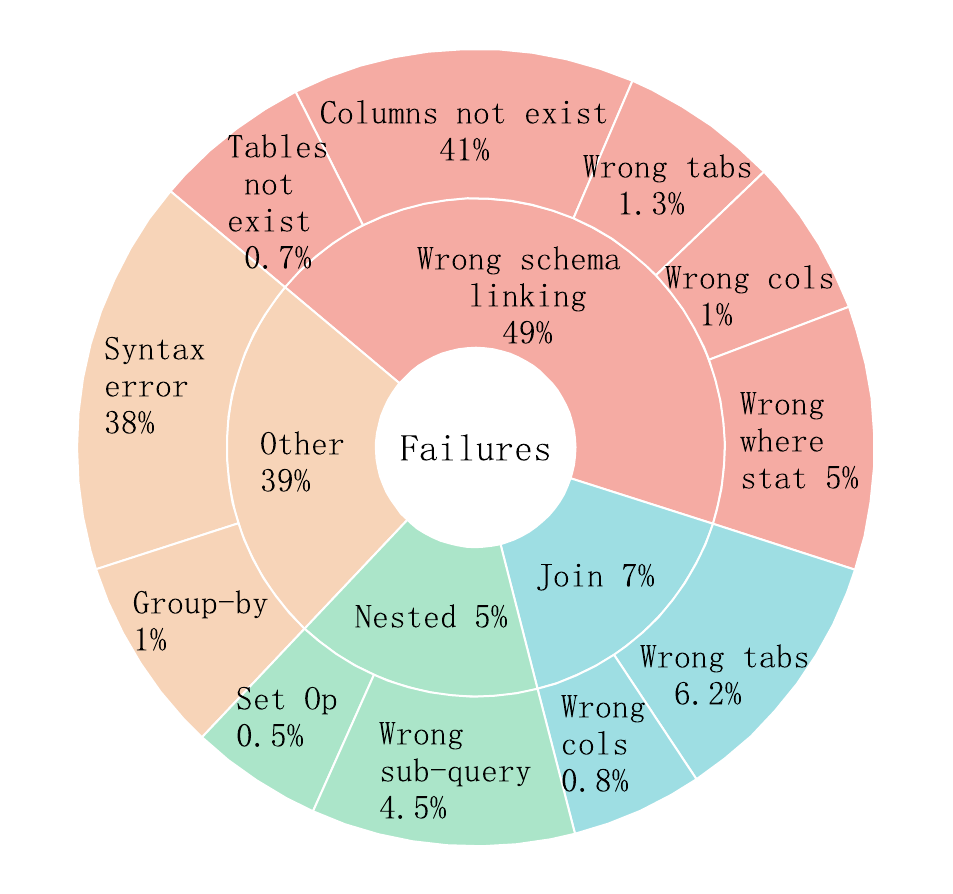}
    \label{fig:error_llama2}}
    \\
    \subfigure[Result of Code Llama.]{
    \includegraphics[width=0.46\textwidth]{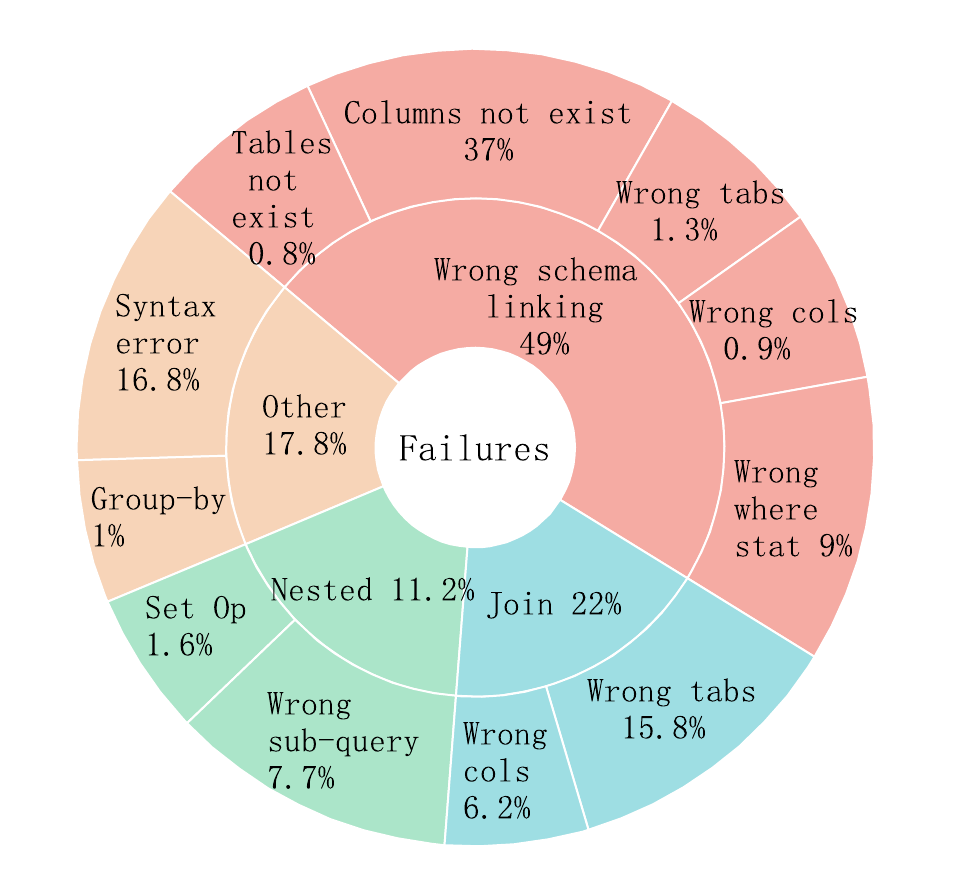}}
    \label{fig:error_Code Llama}
    \caption{Statistics detailing the failures of Llama2 and Code Llama under a zero-shot setting in Text-to-SQL tasks.}
	\label{fig:error_statistics}
\end{figure}

To assess Llama2 and Code Llama in zero-shot Text-to-SQL tasks, we examined queries on BIRD-dev with discrepancies from the gold standard, indicating accuracy failures. The evaluation reveals significant performance shortcomings: Code Llama has an error rate exceeding 90\% on moderate and challenging questions, with an 83\% error rate on simple questions. Llama2 shows an error rate surpassing 98\% across all difficulty levels. A subsequent manual examination categorizes these failures into the following four types.These categories are detailed in Figure~\ref{fig:error_statistics}, with summarized results presented for a clearer understanding.

\begin{enumerate}
    \item \textbf{Wrong schema linking}: The category with the most number of failed queries entails instances where the model struggled to identify column names, table names, and where statements, resulting in inaccurate SQL statements. For instance, in the query ``How many users received commentator badges in 2014'', where the database schema specified a ``badges'' table, the model incorrectly referenced a non-existent ``users'' table.

    \item \textbf{Incorrect JOIN Operation}: This category involves queries requiring a JOIN operation. The model faces difficulties accurately identifying necessary tables or determining correct foreign keys for joins.

    \item \textbf{Inaccurate Nested Structure}: In this category, the gold query employed nesting or set operations, yet the model failed to identify the nested structure accurately or was unable to detect the correct nesting or set operation.

    \item \textbf{Others}: This category encompassed cases that didn't align with any previously defined categories. It featured SQL queries with various issues: syntac error and incorrect GROUP BY.  
\end{enumerate}

In summary, the observed deficiencies in both models, particularly their incapacity to produce valid SQL statements, underscore the urgent need for enhancements in open-source LLMs to adeptly address Text-to-SQL tasks.
\section{Methodology}

In this section, we introduce \ours, a systematic method designed to boost the performance of open-source large language models in Text-to-SQL tasks. In the context of addressing a specific SQL question $\hat{q}$ expressed in natural language within a defined database $\mathcal{D}$, our primary aim is to harness the capabilities of the 
LLMs $\mathcal{M}$ to generate a fitting SQL statement denoted as $s$. This entails the introduction of a dedicated question representation function $q'=\sigma(\hat{q},\mathcal{D})$ designed to enhance the representation of the SQL question $\hat{q}$ by incorporating pertinent details, notably the database schema. Additionally, we introduce a prompt function $q''=\delta(q',\mathcal{Q})$ with the explicit purpose of providing a tailored prompt for the SQL question $q'$, and $\mathcal{Q}$ is the additional information which will be explained in Section~\ref{subsec:incontext}. The final answer is generated through the application of $\mathcal{M}(q'')$.

To enhance the efficacy of this approach, our paper utilizes a synergistic combination of \textbf{question representation}, \textbf{supervised fine-tuning}, and \textbf{in-context learning} techniques. This integrated strategy aims to improve the performance of the language model $\mathcal{M}$ in handling SQL queries denoted by $\hat{q}$. More precisely, the goal of this paper is to maximize the likelihood of LLMs $\mathcal{M}$ generating the correct SQL $s^*$, as outlined below:

\begin{equation*}
\vspace{-1pt}
    \underset{\sigma,\delta,\mathcal{M}}{\max}\qquad \mathbb{P}_{\mathcal{M}}(s^{*}|\delta(\sigma(\hat{q}, \mathcal{D}), \mathcal{Q})),
\end{equation*}
where $\mathbb{P}_{\mathcal{M}}(s^*|\delta(\sigma(\hat{q}, \mathcal{D})))$ is the probability of the model $\mathcal{M}$ generating $s^*$ with the input $\delta(\sigma(\hat{q}, \mathcal{D}))$.

Jointly solving $\sigma$, $\delta$ and $\mathcal{M}$ is challenging. In this paper, we adopt a step-by-step strategy, addressing the problem through \textbf{question representation} to optimize $\sigma$, \textbf{supervised fine-tuning} to optimize $\mathcal{M}$ and \textbf{in-context learning} in inference stage to optimize $\delta$.

\subsection{Question Representation}
\label{subsubsec:question_representation}

\begin{lstlisting}[language=Prompt, caption={Example of \textbf{\openprompt}, where \textbf{DATABASE\_SCHEMA} symbolizes a database schema, \textbf{DATABASE\_DEFN} symbolizes a database definition according to \textbf{DATABASE\_SCHEMA}, \textbf{EXTERNAL\_KNOWLEDGE} represents pertinent external information, and \textbf{TARGET\_QUESTION} denotes the specific question in focus.}, label={lst:openprompt}, float=t]
### Complete sqlite SQL query only and with no explanation 
### SQLite SQL tables are requested to be represented in the following schema.
${DATABASE_SCHEMA}
### Here are SQLite SQL tables, with their properties:
${DATABASE_DEFN}
### Question: ${TARGET_QUESTION}.
### Note that: ${EXTERNAL_KNOWLEDGE}.
SELECT
\end{lstlisting}

\begin{lstlisting}[language=Schema, caption={Definition of full database schema \textbf{DATABASE\_SCHEMA}.}, label={lst:openschema}, float=t]
# TABLE_NAME (
#   COLUMN_NAME:  TYPE, DESCRIPTION, VALUES, PRIMARY_KEY
#   ...
# )
# ...
# FOREIGN KEYS: 
# TABLE_NAME1.COLUMN_NAME1=TABLE_NAME2.COLUMN_NAME2 
# ...
\end{lstlisting}

% The aim of question representation is to optimize the following problem

% \begin{equation}
% \label{def_obj_qr}
% \centering
% \underset{\sigma}{\max}\qquad \mathcal{P}_{\mathcal{M}}(s^{*}|(\delta(\sigma(\hat{q}, \mathcal{D}),\mathcal{Q}))),
% \end{equation}

Gao et al. analyzed five representation methods~\cite{gao2023empowered}, highlighting the effectiveness of \sqlprompt, which introduces ``CREATE TABLE'' SQLs and includes natural language questions in comments to prompt a Language Model (LM)~\cite{chang2023prompt,nan2023enhancing}. In contrast to the Spider benchmark, the Bird benchmark poses challenges for \sqlprompt by including external knowledge evidence such as numeric reasoning, domain knowledge, synonym knowledge, and value illustration.  To address this issue, we introduce \textbf{\openprompt}, a question representation mechanism detailed in Listing~\ref{lst:openprompt}, providing a thorough and nuanced data representation for understanding the database definition. \textbf{\openprompt} comprises four core components: \textbf{DATABASE\_SCHEMA} defines the format of schema; \textbf{DATABASE\_DEFN} represents the concrete schema definition implemented based on the format outlined in \textbf{DATABASE\_SCHEMA}; \textbf{EXTERNAL\_KNOWLEDGE} contains pertinent external information; \textbf{TARGET\_QUESTION} denotes the specific question in focus.

In our initial experimental findings, we underscore the importance of understanding the database definition. To enhance the provision of diverse information for schema linking, we design a \textbf{\openschema} to better bridge the LLMs and database, offering a thorough and nuanced representation of the data. The detailed definition of \textbf{DATABASE\_SCHEMA} is as follows:
 
\begin{itemize}
    \item \textbf{Database definition}: The definitions of the relevant tables are organized sequentially, with each row representing an individual table. Subsequently, foreign key definitions follow, where each row represents a key mapping.
    
    \item \textbf{Table definition}: A table's definition commences with the table name, followed by all column definitions enclosed in brackets. These columns are organized sequentially, with each row representing an individual column.
    
    \item \textbf{Column definition}: The definition of a column starts with the column name, followed by ``:'', and four optional elements: 
    \begin{itemize}
        \item \textbf{TYPE}: Indicates the column type. In the database definition, the corresponding values including ``text,'' ``int,'' ``date,'' ``datetime,'' ``real,'' and ``varchar,'' which align with the definition in SQLite.
        \item \textbf{DESCRIPTION}: Incorporates detailed column information (if available).        
        \item \textbf{VALUES}:         Enumerates categorical values for columns with up to five distinct categories, presented in the format ``(value1, value2, ...)'' for clarity and conciseness.
        \item \textbf{PRIMARY\_KEY}: Indicates whether this column serves as the primary key.
    \end{itemize}
\end{itemize}

\textbf{EXTERNAL\_KNOWLEDGE} in Listing~\ref{lst:openprompt} refers to the additional descriptive information accompanying each query, as provided by the BIRD dataset, that acts as a bridge between human comprehension and the database structure. For example, within the BIRD-DEV dataset, the query ``Please list the phone numbers of the schools with the highest SAT excellence rates in the top 3'' is accompanied by the external knowledge ``Excellence rate = NumGE1500 / NumTstTakr.'' which defines the ``excellence rates''.  In practical applications, this external knowledge can be supplied directly by the users to enhance query understanding.

When dealing with extensive datasets and notably large tables that may exceed the input limitations of open-source  LLMs as described in Listing~\ref{lst:openprompt}, we propose a solution called \textbf{Target Column Truncation}. This approach involves dividing columns $\mathcal{C}$ in the database $\mathcal{D}$ into two subsets: $\mathcal{C}_{\text{target}}$ (containing crucial query columns) and $\overline{\mathcal{C}}_{\text{target}} = \mathcal{C} - \mathcal{C}_{\text{target}}$. By randomly deleting a controlled number of columns in $\overline{\mathcal{C}}_{\text{target}}$, this ensures the question representation remains within the input limits of open-source LLMs.

Moreover, to address the schema linking problem, which involves identifying references to database schema elements and condition values in natural language queries, enhancing generalizability and supporting complex query synthesis~\cite{lei2020re} (identified as the second significant failure detailed in  Figure~\ref{fig:error_statistics}), we introduce two Chain-of-Thought (\textbf{COT}) templates for schema linking in this paper. The initial simple template (\textbf{COT-SP}), explained in Listing~\ref{lst:col_simple} (Appendix~\ref{app:promt_cot}), takes a direct approach, guiding the Language Model (LLM) to explore related tables and columns before formulating the complete SQL script. The second template, the skeleton-based Chain-of-Thought (\textbf{COT-SK}), detailed in Listing~\ref{lst:col_skeleton} (Appendix~\ref{app:promt_cot}), refines \textbf{COT-SP} by introducing a skeleton generation phase before the actual SQL script generation. This ``skeleton'' serves as the basic query framework without specific values or conditions. In our methodology, we preserve SQL keywords in their original form, replacing other components with a designated placeholder symbol, typically represented as ``\_''. For instance, the SQL query ``SELECT movie\_title FROM movies WHERE movie\_release\_year = 1945 ORDER BY movie\_popularity DESC LIMIT 1'' corresponds to the skeleton ``SELECT \_ FROM \_ WHERE \_ ORDER BY \_ DESC LIMIT \_''.

\subsection{Supervised Fine-Tuning for Text-to-SQL}

% For Text-to-SQL, given a large language model $\mathcal{M}$, a set of Text-to-SQL training data $\mathcal{T}=\{(\mathcal{D}_i, \hat{q_i}, s^{*}_i)\}$, where $\hat{q_i}$ and $s^{*}_i$ are the natural language question and its golden query on database $\mathcal{D}_i$, the objective of supervised fine-tuning is to minimize the following empirical loss:
% \begin{equation*}
%     \min_{\mathcal{M}} \sum_{i=1}^{|\mathcal{T}|}{\mathcal{L}(\mathcal{M}(\delta(\sigma(\hat{q_i}, \mathcal{D}_i),\mathcal{Q})), s^{*}_i)},
% \end{equation*}

In the general domain, each entry in the supervised data set \(\mathcal{T} = \{(p_i, s^{*}_i)\}\) comprises a prompted input \(p_i = \sigma(\delta(\hat{q_i}, \mathcal{D}_i))\) with the natural language question $\hat{q_i}$ and a golden query on database $\mathcal{D}_i$. The prompted input is constructed by applying a transformation \(\sigma\) to the composition \(\delta(\hat{q_i}, \mathcal{D}_i)\). Subsequently, LoRA~\cite{hu2021lora} is employed for fine-tuning the given LLMs. Following the fine-tuning process, the optimized language model \(\mathcal{M}^*\) becomes capable of performing inference tasks. In this context, inference involves prompting \(\mathcal{M}^*\) with natural language questions to generate corresponding queries. It's noteworthy that the same question representation \(\sigma\) and prompt \(\delta\) are employed consistently throughout both the fine-tuning and inference phases.

\subsection{Few-shot Learning for Text-to-SQL}
\label{subsec:incontext}

% \begin{equation}
% \underset{\delta}{\max}\qquad \mathcal{P}_{\mathcal{M}}(s^{*}|(\delta(\sigma(\hat{q}, \mathcal{D})),\mathcal{Q})),
% \end{equation}

We leverage few-shot learning to enhance the Text-to-SQL performance of open-source LLMs by incorporating a limited number of examples in the input prompts. This method allows for more effective training, improving the overall system performance. Suppose we have an additional sample set $\mathcal{T}^{a}$ with the same format as $\mathcal{T}$. We aim to select a set of $k$ triples $\mathcal{Q} \subset \mathcal{T}^{a}$ to maximize the probability of LLMs $\mathcal{M}$ generating the correct SQL $s^*$ for the target question $\hat{q}$ and database $\mathcal{D}$. When $k=0$, few-shot learning reverts to zero-shot learning. This study focuses on \emph{cross-domain Text-to-SQL}, where the target database $\mathcal{D}$ is distinct from those encompassed in $\mathcal{Q}$, i.e., $\mathcal{D} \notin \mathcal{D}_i$ holds for all $(\mathcal{D}_i, \hat{q_i}, s^{*}_i) \in \mathcal{Q}$.

% \begin{lstlisting}[language=Prompt, label={lst:org_os}, caption={Example of \fiorg.}, float=t]
% /* Given the following database schema: */
% ${DATABASE_SCHEMA}
% /* Answer the following: How many authors are there? */
% SELECT count(*) FROM authors

% /* Given the following database schema: */
% ${REDUCED_DATABASE_SCHEMA}
% /* Answer the following: How many farms are there? */
% SELECT count(*) FROM farm

% ${TARGET_QUESTION}
% \end{lstlisting}

In this research, we introduce a novel methodology named \textbf{\openexample}, outlined in Listing~\ref{lst:prompt_few_shot}(Appendix~\ref{app:promt_shot}), as a systematic approach for the selection and organization of $\mathcal{Q}$. When it comes to example selection, the conventional strategy involves presenting both example questions and their corresponding SQL queries while excluding the database schema to manage token costs. However, this exclusion of the database schema may have a detrimental impact on performance. To address this concern, we propose presenting example questions, corresponding database schema, and SQL queries together. For a given question $\hat{q}$ with database $\hat{\mathcal{D}}$ and each $i$-th training example $(\mathcal{D}_i, q_i, s_i)\in\mathcal{T}^{a}$, initial step involves computing the question similarity between $e_{\mathcal{M}}(\hat{q})$ and $e_{\mathcal{M}}(q_{i})$, the database similarity between $\gamma^{d}_{i}=e_{\mathcal{M}}(\hat{\mathcal{D}})$ and $e_{\mathcal{M}}(\mathcal{D}_{i})$, and the SQL similarity between $\gamma^{s}_{i}=e_{\mathcal{M}}(\mathcal{M}(\delta(\sigma(\hat{q}),\emptyset)))$ and $e_{\mathcal{M}}(s_{i})$, where $e_{\mathcal{M}}(.)$ returns the embedding of the input by $\mathcal{M}$ and cosine similarity is employed. Subsequently, we compute the average similarity $\gamma^{a}_{i}=\frac{1}{3}(\gamma^{q}_{i}+\gamma^{d}_{i}+\gamma^{s}_{i})$. Finally, we select $k$ examples with the highest average similarities from $\{\gamma^{a}_{1},\cdots,\gamma^{a}_{|\mathcal{T}^{a}|}\}$ to form $\mathcal{Q}$.

In the domain of example organization, diverse strategies are employed. Some maintain both the question-SQL mapping and the database schema, while others prioritize the former, excluding the token-costly database schema. To efficiently preserve both mappings simultaneously, we propose the \textbf{Example Column Truncation} strategy. This method selectively removes unnecessary columns in the example database based on the dissimilarity of a table to the target database. Let $\gamma^{a}_{0}=1$ denote the similarity of the question $\hat{q}$ with itself, and $\{\gamma^{a}_{1},\cdots,\gamma^{a}_{k}\}$ represent the similarities of question $\hat{q}$ with the selected $k$ examples in $\mathcal{Q}$. Truncation rates, $\{\rho_{0},\cdots,\rho_{k}\}$, are calculated using the softmax function with a temperature parameter $T$ (typically set to $1$) applied to $\{1/\gamma^{a}_{0}, \cdots, 1/\gamma^{a}_{k}\}$. It is crucial to note that adjusting $T$ within the softmax function enables fine-tuning of the distribution of truncation rates. Columns can then be removed according to these truncation rates. Ultimately, these truncated examples can be ordered in ascending order based on their similarities to the target example and subsequently integrated into the prompt for in-context learning.

\section{Experiment}
\label{sec:exp}
In this section, we first introduce our experimental settings. 
Then we conduct extensive comparisons with existing solutions in the original LLM models, supervised fine-tuned LLM models without Chain-of-Thought and supervised fine-tuned LLM models with Chain-of-Thought respectively. 
After that, we further compare them in terms of token efficiency to inspire more efficient solutions.

\subsection{Setting}
\label{sec:setting}

\textbf{Dataset:} Our evaluation utilizes the BIRD dataset~\cite{DBLP:journals/corr/abs-2305-03111}, as detailed in Section~\ref{sec:error_analysis}. Specifically, we employ the development split, referred to as \emph{BIRD-dev}, since the test split has not been publicly released. In cases involving few-shot scenarios, we make use of the training split of BIRD to provide example candidates for evaluation.

\noindent\textbf{Metrics}. To ensure a fair comparison, we adopt the execution accuracy (\textbf{EX}) metric~\cite{DBLP:journals/corr/abs-2305-03111}, which assesses the agreement in execution output between the predicted SQL query and the ground truth SQL query across various database instances.

\noindent\textbf{Implementation details:} In exploring the capabilities of open-source LLMs, we selected \textbf{\lmc}-7B~\cite{llama2} and \textbf{\clm}-7B~\cite{llama2}, both equipped with 7 billion parameters. Our experiments were conducted on a NVIDIA A100 80 G GPU for consistent comparison across all methods. To address memory limitations in supervised fine-tuning, we set the maximal context length to 2048 for both \lmc and \clm. During evaluation, we allocated $200$ tokens for response generation, and, by default, we set the argument temperature to $0.001$ to eliminate the influence of randomness. For post-processing, we adhered to established practices, extracting the first SQL query in the response and removing any additional output.
\begin{figure*}[t]
    \centering
    \subfigure[Results of GPT-3.5-TURBO.]{
    \includegraphics[width=0.98\textwidth]{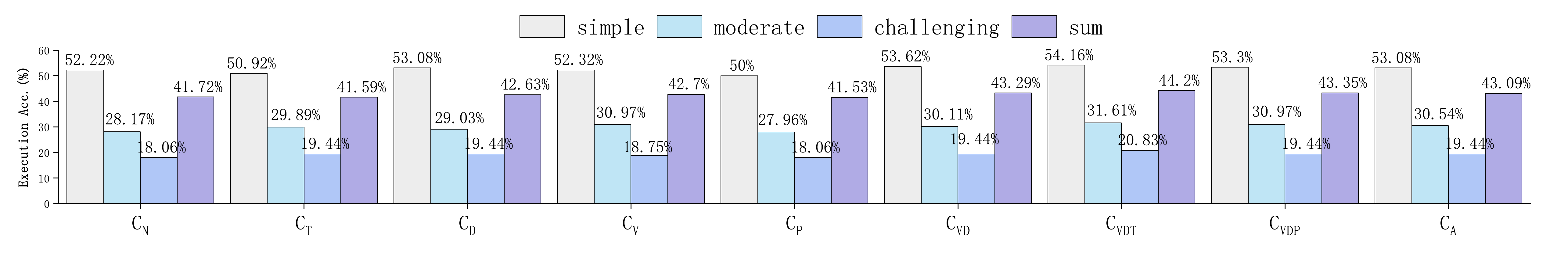}
    \label{fig:qr_gpt}}
    \\
    \subfigure[Results of Llama 2.]{
    \includegraphics[width=0.98\textwidth]{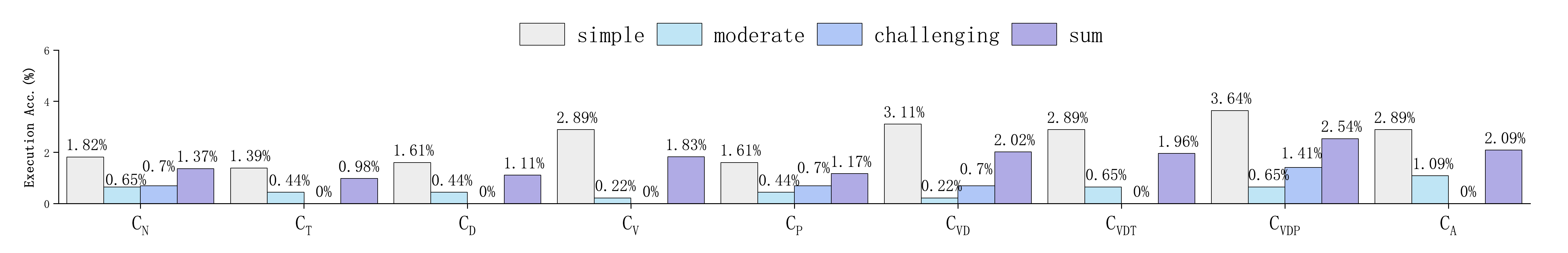}
    \label{fig:qr_llama2}}
    \\
    \subfigure[Results of Code Llama.]{
    \includegraphics[width=0.98\textwidth]{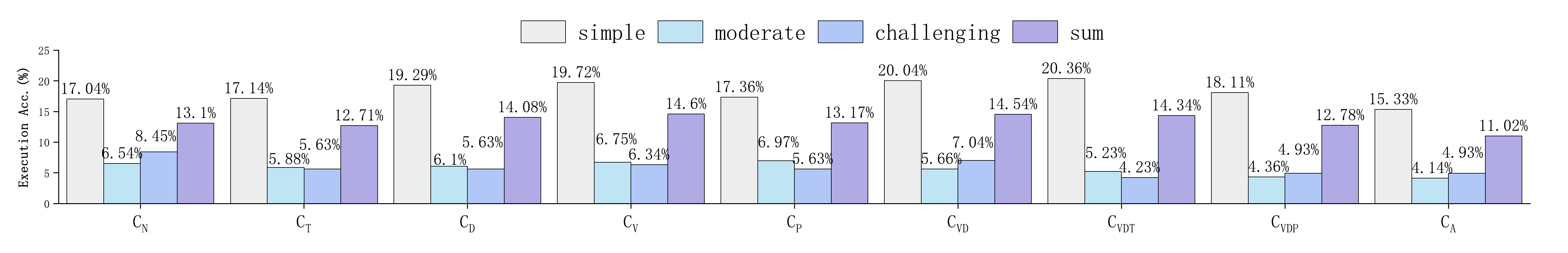}
    \label{fig:qr_codellama}}
    \caption{The comparative results of GPT-3.5-TURBO, Llama2, and Code Llama across different table schemas on BIRD-dev.}
	\label{fig:qr_zero}
\end{figure*}

\subsection{In-Context Learning without Fine-tuning}

In this subsection, we conduct a comprehensive evaluation of both closed-source LLMs and open-source LLMs, specifically employing GPT-3.5-TURBO in conjunction with two distinct open-source Language Models: \lmc\ and \clm. This evaluation encompasses a variety of question representation, example selection, and organization strategies.

\noindent\textbf{Comparative Analysis with Question Representations}.

Note that the table definition becomes more informative with the inclusion of optional column elements. We compare nine types of column definitions, each incorporating various optional column elements: $C_{N}$ with none of the optional column elements; $C_{T}$ with \textbf{TYPE}; $C_{D}$ with \textbf{DESCRIPTION}; $C_{V}$ with \textbf{VALUES}; $C_{P}$ with \textbf{PRIMARY\_KEY}; $C_{VD}$ with \textbf{DESCRIPTION} and \textbf{VALUES}; $C_{VDT}$ with \textbf{DESCRIPTION}, \textbf{VALUES}, and \textbf{TYPE}; $C_{VDP}$ with \textbf{DESCRIPTION}, \textbf{VALUES}, and \textbf{PRIMARY\_KEY}; $C_{A}$ with \textbf{TYPE}, \textbf{DESCRIPTION}, \textbf{VALUES}, and \textbf{PRIMARY\_KEY}. The detailed definitions of these nine types of column definitions are provided in Appendix~\ref{app:prompt:columns}.

% \begin{itemize}
%     \item $C_{N}$: With none of the optional column elements.
%     \item $C_{T}$: With \textbf{TYPE}.
%     \item $C_{D}$: With \textbf{DESCRIPTION}.
%     \item $C_{V}$: With \textbf{VALUES}.
%     \item $C_{P}$: With \textbf{PRIMARY\_KEY}.
%     \item $C_{VD}$: With \textbf{DESCRIPTION} and \textbf{VALUES}.
%     \item $C_{VDT}$: With \textbf{DESCRIPTION}, \textbf{VALUES}, and \textbf{TYPE}.
%     \item $C_{VDP}$: With \textbf{DESCRIPTION}, \textbf{VALUES}, and \textbf{PRIMARY\_KEY}.
%     \item $C_{A}$: With \textbf{TYPE}, \textbf{DESCRIPTION}, \textbf{VALUES}, and \textbf{PRIMARY\_KEY}.
% \end{itemize}

In Figure~\ref{fig:qr_zero} , we present a comparative analysis of GPT-3.5-TURBO, Llama2, and Code Llama across various table schemas on BIRD-dev. Examining GPT-3.5-TURBO's performance in various column definitions (Figure~\ref{fig:qr_gpt}), we note that incorporating column values has the most substantial positive impact, and adding column descriptions also contributes to performance improvement. Conversely, including data type and primary key appears to have a negative effect, particularly when introducing only one optional column. The combined inclusion of both column values and descriptions leads to further performance improvement, and the introduction of column type and primary key enhances performance, with the former contributing the most significant improvement.

Similar trends are evident for Llama2 and Code Llama. Given the notably subpar performance of Llama2 and Code Llama, the insights gained from GPT-3.5-TURBO's results are more meaningful and can inform enhancements for Llama2 and Code Llama. In summary, optimal performance is attained by $C_{VDT}$ which introduces column values, descriptions, and types for enhanced results, yielding an improvement from 37.72\%~\footnote{https://bird-bench.github.io/} to 44.2\% for GPT-3.5-TURBO. However, $C_{V}$ which only introduces column values for token-saving purposes results in a slight decrease, reaching 42.7\% for GPT-3.5-TURBO. Hence, the choice between $C_{VDT}$ and $C_{V}$ depends on specific purposes.

\noindent\textbf{Comparative Analysis with COT}.

\begin{figure}[t]
    \centering
    \subfigure[Result of Llama 2.]{
    \includegraphics[width=0.49\textwidth]{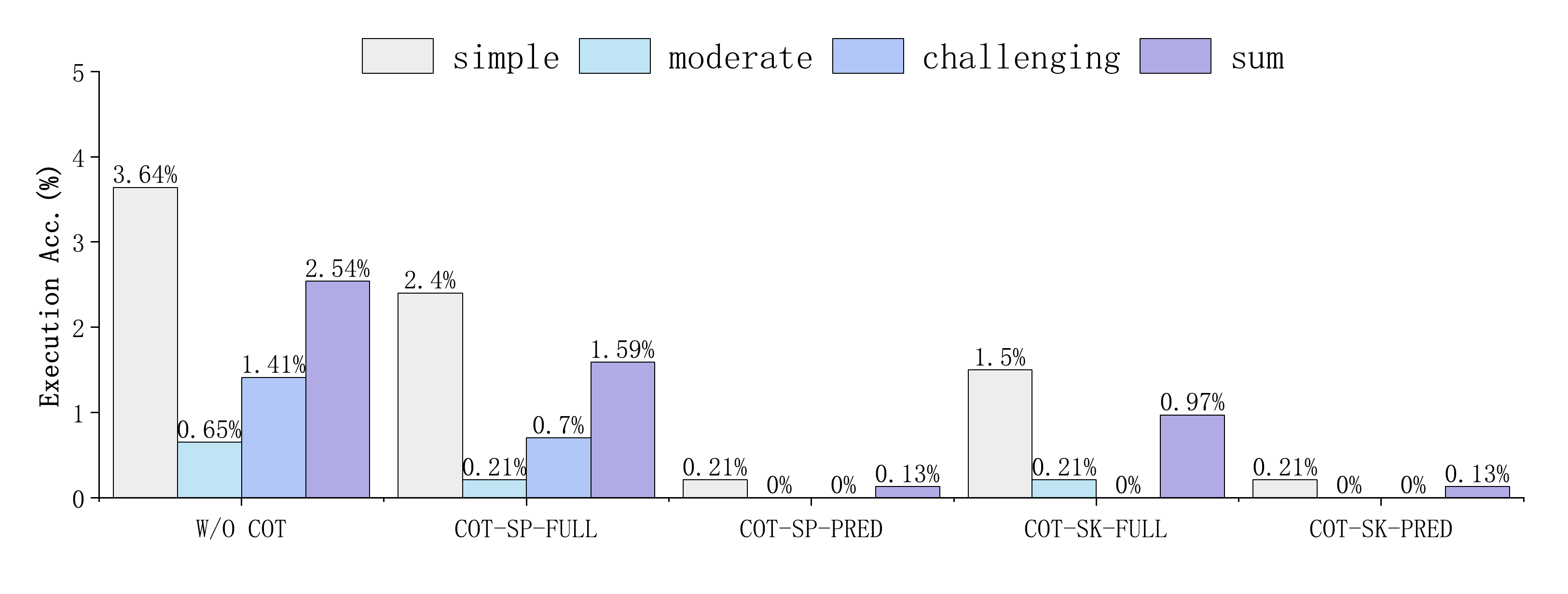}
    \label{fig:llama2_cot}}
    \\
    \subfigure[Result of Code Llama.]{
    \includegraphics[width=0.49\textwidth]{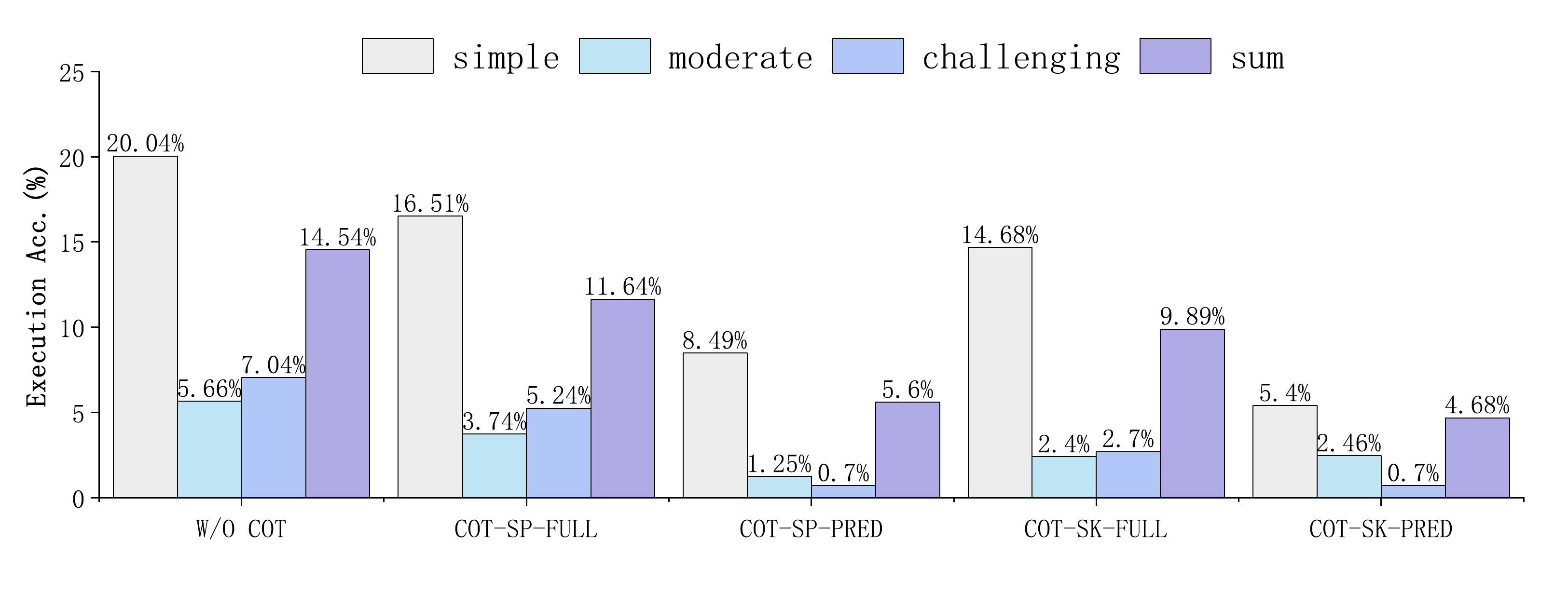}}
    \label{fig:codellama_cot}
    \caption{Zero-shot outcomes of original open-source LLMs on the BIRD-dev dataset. ``W/O COT'' indicates inference without Chain-of-Thought (COT), while ``COT-SP-PRED'' and ``COT-SK-PRED'' represent inference with the COT-SP and COT-SK methods, respectively, using only the definition of predicted tables and columns. Additionally, ``COT-SP-FULL'' and ``COT-SK-FULL'' represent inference with the COT-SP and COT-SK methods, respectively, incorporating the complete database definition.
    }
	\label{fig:qr_cot_zero}
\end{figure}

In Figure~\ref{fig:qr_cot_zero}, we provide a comprehensive comparison of zero-shot outcomes by Llama2 and Code Llama across five methods on BIRD-dev. In summary, the introduction of COT results in performance deterioration, highlighting a challenge for open-source LLMs in executing COT for Text-to-SQL tasks. Notably, ``COT-SP-FULL'' and ``COT-SK-FULL'' outperform ``COT-SP-PRED'' and ``COT-SK-PRED'', suggesting that inaccuracies in predicting tables and columns may negatively impact SQL generation.

% However, it should be noted that incorporating the entire database definition in the final SQL generation step may significantly increase the token cost in the final stage.

\noindent\textbf{Comparative Analysis of Few-shot Learning}.

\begin{table}[h]
\caption{Few-shot evaluation with original Llama2.}
\label{tab:llama2_ft}
\resizebox{0.5\textwidth}{!}{
    \begin{tabular}{llllll}
    \hline
    Few-shot                                    & Selection                  & simple & moderate & challenging & sum \\ \hline
    0-shot                                      &                            & 2.89    & 0.65      & 0.0         & 1.96 \\ \hline
    \multicolumn{1}{c}{\multirow{4}{*}{1-shot}} & Random                     & 9.43    & 2.83      & 0.7         & 6.65 \\
    \multicolumn{1}{c}{}                        & Q similarity        & 7.93    & 0.87      & 2.11         & 5.28 \\
    \multicolumn{1}{c}{}                        & Q+S similarity & 9.6    & 1.9      & 0.7         & 6.5 \\
    \multicolumn{1}{c}{}                        & DAIL-SQL~\cite{gao2023empowered} & 7.72    & 1.53      & 1.41         & 5.28 \\
    \multicolumn{1}{c}{}                        & Ours             & 10.61   & 1.53      & 1.41         & 7.04 \\ \hline
    \multirow{4}{*}{3-shot}                     & Random                     & 2.57    & 1.09      & 0.0         & 1.89 \\
    & Q similarity        & 2.24    & 0.44      & 0.70         & 1.55 \\
    & Q+S similarity & 2.04    & 0.44     & 0.70         & 1.43 \\
    \multicolumn{1}{c}{}                        & DAIL-SQL~\cite{gao2023empowered} & 4.61   & 0.65      & 0.7         & 3.06 \\
    & Ours             & 6.43   & 0.65      & 0.0         & 4.11 \\ \hline
    \end{tabular}
}
\end{table}

\begin{table}[h]
\caption{Few-shot evaluation with original Code Llama.}
\label{tab:codellama_ft}
\resizebox{0.5\textwidth}{!}{
    \begin{tabular}{llllll}
    \hline
    Few-shot                                    & Selection                  & simple & moderate & challenging & sum  \\ \hline
    0-shot                                      &                            & 20.36   & 5.23      & 4.23         & 14.34 \\ \hline
    \multicolumn{1}{c}{\multirow{4}{*}{1-shot}} & Random                     & 25.19   & 11.33      & 9.15         & 19.56 \\
    \multicolumn{1}{c}{}                        & Q similarity        & 26.47   & 9.80     & 11.27         & 20.08 \\
    \multicolumn{1}{c}{}                        & Q+S similarity & 24.33   & 10.68      & 8.45         & 18.77 \\
    \multicolumn{1}{c}{}                        & DAIL-SQL~\cite{gao2023empowered} & 27.22   & 12.42      & 9.86         & 21.19 \\
    \multicolumn{1}{c}{}                        & Ours             & 28.72   & 11.33      & 11.27         & 21.90 \\ \hline
    \multirow{4}{*}{3-shot}                     & Random                     & 24.22   & 10.46      & 6.34         & 18.45 \\
    & Q similarity        & 26.15   & 12.64      & 7.75        & 20.40 \\
    & Q+S similarity & 25.51   & 11.11      & 6.34         & 19.43 \\
    \multicolumn{1}{c}{}                        & DAIL-SQL~\cite{gao2023empowered} & 30.01   & 12.63      & 7.04         & 22.68 \\
    & Ours            & 30.23   & 14.60      & 7.75         & 23.47 \\ \hline
    \end{tabular}
}
\end{table}

In our experimentation to assess the few-shot learning capabilities of open-source LLMs such as Llama2 and Code Llama, detailed in Tables~\ref{tab:llama2_ft} and~\ref{tab:codellama_ft}, examples were chosen using the $C_{VDT}$ column definition through different methods: Random selection, Q (Question) similarity, Q+S (Question and Schema) similarity, DAIL-SQL~\cite{gao2023empowered} (Question and pre-generated SQL), and our proposed Open Example Curation method. Notably, our proposed approach surpassed other example selection methods, showcasing superior performance. Additionally, we observed a positive impact on performance with an increase in the number of examples. Specifically, our method demonstrated a substantial improvement of approximately $260\%$ with the integration of one example into Llama2 and $53\%$ with one example into Code Llama. However, introducing three examples led to a decline in performance for Llama2 but a slight increase for Code Llama.
\subsection{Supervised Fine-tuning without COT}

In this subsection, our exploration centers on supervised fine-tuning in Text-to-SQL, specifically without COT .

\noindent\textbf{Comparative Analysis with Question Representations}.

\begin{figure*}[t]
    \centering
    \subfigure[Results of supervised-finetuned Llama 2.]{
    \includegraphics[width=0.98\textwidth]{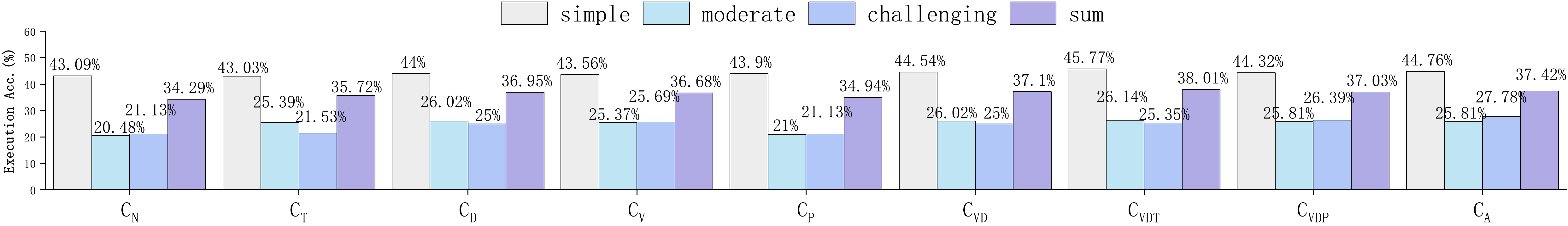}
    \label{fig:qr_llama2_sft}}
    \\
    \subfigure[Results of supervised-finetuned Code Llama.]{
    \includegraphics[width=0.98\textwidth]{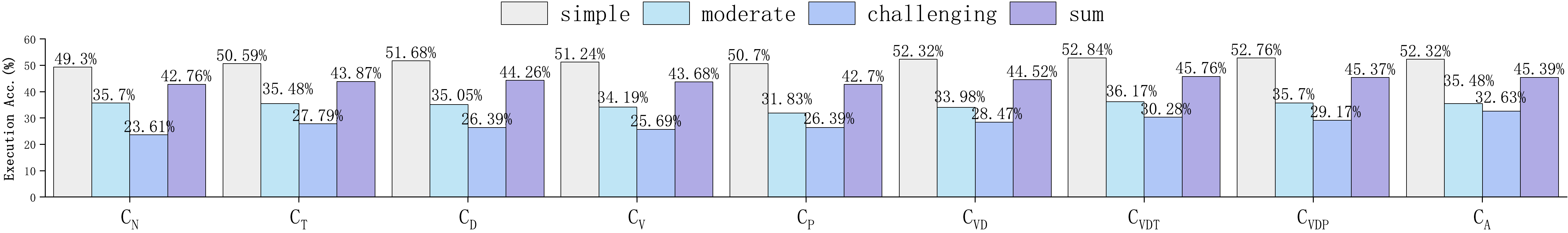}
    \label{fig:qr_codellama_sft}}
    \caption{The comparative results of supervised-finetuned Llama2 and Code Llama across various table schemas on BIRD-dev.}
	\label{fig:qr_sft}
\end{figure*}

Figure~\ref{fig:qr_sft} present a comparative analysis of the results obtained by Llama2 and Code Llama after supervised fine-tuning across various table schemas on BIRD-dev. Notably, our initial observations reveal performance improvements through supervised fine-tuning for both Llama2 and Code Llama, irrespective of the table schema used.

The outcomes depicted in Figures~\ref{fig:qr_llama2_sft} and \ref{fig:qr_codellama_sft} align with the trends observed in Figures~\ref{fig:qr_llama2} and \ref{fig:qr_codellama}, consistently indicating that supervised fine-tuned Code Llama consistently outperforms fine-tuned Llama2 in Text-to-SQL tasks. Notably, the inclusion of column description emerges as the most impactful improvement, particularly when considering individual column elements. However, the most substantial positive impact is realized when introducing column values, descriptions, and types together, as corroborated by the findings in Figure~\ref{fig:qr_gpt}. This observation underscores the notion that the performance of both fine-tuned Code Llama and Llama2 closely mirrors that of GPT-3.5-TURBO, showcasing heightened proficiency in Text-to-SQL tasks.

Finally, our analysis reveals a noteworthy increase from $2.54\%$ to $38.01\%$ for Llama2 and an increase from $14.54\%$ to $45.76\%$ for Code Llama, underscoring the significant potential of supervised fine-tuning in enhancing Text-to-SQL performance.

\noindent\textbf{Comparative Analysis of Few-shot Learning}.

\begin{table}[h]
\caption{Few-shot evaluation with Supervised Fine-tuned Llama2.}
\label{tab:llama2_ft_sft}
\resizebox{0.5\textwidth}{!}{
    \begin{tabular}{llllll}
    \hline
    Few-shot                                    & Selection                  & simple & moderate & challenging & sum \\ \hline
    0-shot                                      &                            & 45.77    & 26.16      & 25.35         & 38.01 \\ \hline
    \multicolumn{1}{c}{\multirow{4}{*}{1-shot}} & Random                     & 44.8    & 23.52      & 18.3         & 35.98 \\
    \multicolumn{1}{c}{}                        & Q similarity        & 42.12    & 22      & 21.83         & 34.22 \\
    \multicolumn{1}{c}{}                        & Q+S similarity & 43.73    & 26.14      & 25.35         & 36.77 \\
    \multicolumn{1}{c}{}                        & DAIL-SQL~\cite{gao2023empowered} & 42.12    & 21.79      & 22.54         & 34.22 \\
    \multicolumn{1}{c}{}                        & Ours             & 40.73   & 23.09      & 23.94         & 33.9 \\ \hline
    \multirow{4}{*}{3-shot}                     & Random                     & 35.58   & 19.39      & 17.61        & 29.07 \\
    & Q similarity        & 36.01    & 20.7      & 21.83         & 30.12 \\
    & Q+S similarity & 40.41    & 21.57      & 21.13         & 32.99 \\
    \multicolumn{1}{c}{}                        & DAIL-SQL~\cite{gao2023empowered} & 33.12    & 25.33      & 23.94         & 29.94 \\
    & Ours             & 38.05   & 21.35      & 22.54         & 31.62 \\ \hline
    \end{tabular}
}
\end{table}

\begin{table}[h]
\caption{Few-shot evaluation with Supervised Fine-tuned Code Llama.}
\label{tab:codellama_ft_sft}
\resizebox{0.5\textwidth}{!}{
    \begin{tabular}{llllll}
    \hline
    Few-shot                                    & Selection                  & simple & moderate & challenging & sum  \\ \hline
    0-shot                                      &                            & 52.84   & 36.17      & 30.28         & 45.76 \\ \hline
    \multicolumn{1}{c}{\multirow{4}{*}{1-shot}} & Random                     & 46.62   & 31.81      & 26.76         & 40.35 \\
    \multicolumn{1}{c}{}                        & Q similarity        & 43.94   & 27.89      & 28.87         & 37.74 \\
    \multicolumn{1}{c}{}                        & Q+S similarity & 44.80   & 30.5      & 26.06         & 38.79 \\
    \multicolumn{1}{c}{}                        & DAIL-SQL~\cite{gao2023empowered} & 43.84   & 28.98      & 26.06         & 37.74 \\
    \multicolumn{1}{c}{}                        & Ours             & 44.59   & 28.32      & 23.94         & 37.81 \\ \hline
    \multirow{4}{*}{3-shot}                     & Random                     & 39.23   & 26.8      & 22.54         & 33.96 \\
    & Q similarity        & 37.94   & 25.71      & 15.49         & 32.2 \\
    & Q+S similarity & 42.44   & 26.14      & 24.65         & 35.92 \\
    \multicolumn{1}{c}{}                        & DAIL-SQL~\cite{gao2023empowered} & 37.19  & 27.02     & 31.69         & 33.64 \\
    & Ours            & 40.84   & 24.62      & 22.54         & 34.29 \\ \hline
    \end{tabular}
}
\end{table}

We conducted additional experiments to assess the few-shot learning capabilities of Llama2 and Code Llama after supervised fine-tuning, using examples from BIRD-DEV with the $C_{VDT}$ column definition (Tables~\ref{tab:llama2_ft_sft} and~\ref{tab:codellama_ft_sft}). In contrast to prior results (Tables~\ref{tab:llama2_ft} and~\ref{tab:codellama_ft}), both models exhibited a decline in performance with additional examples. This might be attributed to the models becoming overly focused on the zero-shot prompt, hindering their understanding of test prompts with extra examples. Addressing this issue is crucial for future work to enhance the few-shot capabilities of open-source LLMs after supervised fine-tuning.

% The examples were selected using various methods, including Random, Q (Question) similarity, Q+S (Question and Schema) similarity, DAIL-SQL~\cite{gao2023empowered} (Question and pre-generated SQL), and our proposed Open Example Curation method.

\subsection{Supervised Fine-tuning with COT}
\label{sec:exp_sft_cot}

Within this subsection, our exploration centers on assessing the potential for enhancing supervised fine-tuning performance in open-source LLMs through the incorporation of COT.

\begin{figure}[h]
    \centering
    \subfigure[Result of Llama 2.]{
    \includegraphics[width=0.46\textwidth]{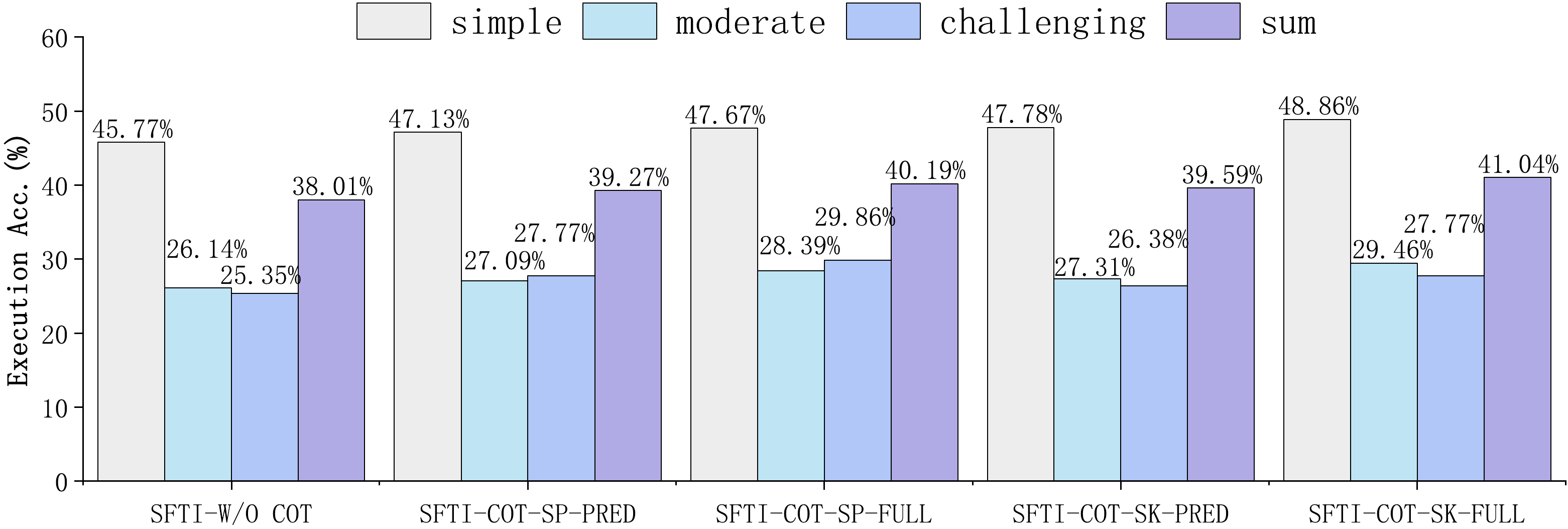}
    \label{fig:llama2_cot_sft}}
    \\
    \subfigure[Result of Code Llama.]{
    \includegraphics[width=0.46\textwidth]{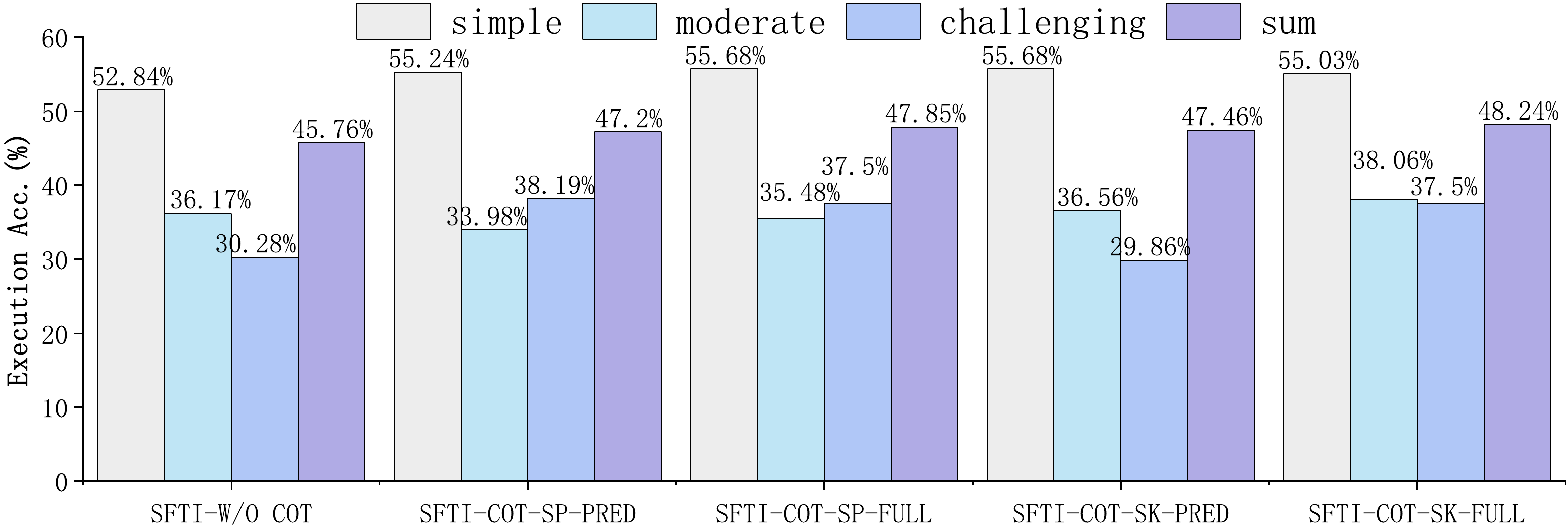}}
    \label{fig:codellama_cot_sft}
    \caption{Zero-shot outcomes of open-source Large Language Models (LLMs) following supervised fine-tuning, both with and without Chain-of-Thought, on the BIRD-dev dataset. ``SFTI-W/O COT'' indicates supervised fine-tuning and inference without Chain-of-Thought (COT), while ``SFTI-COT-SP-PRED'' and ``SFTI-COT-SK-PRED'' represent supervised fine-tuning and inference with the COT-SP and COT-SK methods, respectively, using only the definition of predicted tables and columns. Additionally, ``SFTI-COT-SP-FULL'' and ``SFTI-COT-SK-FULL'' represent supervised fine-tuning and inference with the COT-SP and COT-SK methods, respectively, incorporating the complete database definition.}
	\label{fig:cot_sft}
\end{figure}

\noindent\textbf{Comparative Analysis of COT}.

In Figure~\ref{fig:cot_sft}, we provide a comprehensive comparison of zero-shot outcomes by Llama2 and Code Llama across five methods on BIRD-dev, where the prefix ``SFTI-'' denotes supervised fine-tuning and inference. 
Both ``SFTI-COT-SP-FULL'' and ``SFTI-COT-SK-FULL'' outperform ``SFTI-COT-SP-PRED'' and ``SFTI-COT-SK-PRED'', indicating that introducing the target database schema improves step-by-step inference performance. In contrast to the results in Figure~\ref{fig:qr_cot_zero}, ``SFTI-COT-SK-FULL'' surpasses ``SFTI-COT-SP-FULL'', revealing the emergence of skeleton generation capability in Llama2 and Code Llama after fine-tuning. Compared to the results in Figure~\ref{fig:qr_sft}, ``SFTI-COT-SK-FULL'' boosts Llama2 from 37.42\% to 41.04\% and Code Llama from 45.39\% to 48.24\%.

\noindent\textbf{Comparative Analysis of Few-shot Learning}.

\begin{table}[h]
\caption{Few-shot evaluation of Llama2 with the ``SFTI-COT-SK-FULL'' method.}
\label{tab:llama2_ft_sft_cot}
\resizebox{0.5\textwidth}{!}{
    \begin{tabular}{llllll}
    \hline
    Few-shot                                    & Selection                  & simple & moderate & challenging & sum \\ \hline
    0-shot                                      &                            & 48.86    & 29.46      & 27.77         & 41.04 \\ \hline
    \multicolumn{1}{c}{\multirow{4}{*}{1-shot}} & Random                     & 42.92   & 23.52      & 20.6         & 34.84 \\
    \multicolumn{1}{c}{}                        & Q similarity        & 42.05      & 22.53      & 21.6         & 34.31 \\
    \multicolumn{1}{c}{}                        & Q+S similarity & 43.17    & 25.05      & 22.76         & 35.84 \\
    \multicolumn{1}{c}{}                        & DAIL-SQL~\cite{gao2023empowered} & 42.57    & 21.16      & 20.42         & 34.11 \\
    \multicolumn{1}{c}{}                        & Ours             & 43.41   & 23.09      & 21.83         & 35.33 \\ \hline
    \multirow{4}{*}{3-shot}                     & Random                     & 35.94    & 18.3      &  17.6        & 28.96 \\
    & Q similarity        & 36.01    & 18.7      & 19.71       & 29.32 \\
    & Q+S similarity & 34.8   & 19.1      & 20.42         & 28.79 \\
    \multicolumn{1}{c}{}                        & DAIL-SQL~\cite{gao2023empowered} & 35.04    & 21.79      & 21.12        & 29.97 \\
    & Ours          & 38.48   & 22.0      & 19.71         & 31.8 \\ \hline
    \end{tabular}
}
\end{table}

\begin{table}[h]
\caption{Few-shot evaluation of Code Llama with the ``SFTI-COT-SK-FULL'' method.}
\label{tab:codellama_ft_sft_cot}
\resizebox{0.5\textwidth}{!}{
    \begin{tabular}{llllll}
    \hline
    Few-shot                                    & Selection                  & simple & moderate & challenging & sum  \\ \hline
    0-shot                                      &         & 55.03   & 38.06      & 37.5         & 48.24  \\ \hline
    \multicolumn{1}{c}{\multirow{4}{*}{1-shot}} & Random       & 45.64   & 27.88      & 23.94         & 38.31 \\
    \multicolumn{1}{c}{}                        & Q similarity        & 43.94   & 26.14      & 24.6         & 36.82 \\
    \multicolumn{1}{c}{}                        & Q+S similarity & 44.16    & 26.58      &  23.94        & 37.02 \\
    \multicolumn{1}{c}{}                        & DAIL-SQL~\cite{gao2023empowered} & 43.94   & 26.14      & 23.94        & 36.76 \\
    \multicolumn{1}{c}{}                        & Ours             & 45.33   & 26.57      & 24.64         & 37.8 \\ \hline
    
    \multirow{4}{*}{3-shot}                     & Random                     & 39.44   & 22.22      & 16.2         & 32.13 \\
    & Q similarity        & 39.22   & 22.65      & 16.9         & 31.55 \\
    & Q+S similarity & 39.87   & 22.22      & 18.3        & 32.59 \\
    \multicolumn{1}{c}{}                        & DAIL-SQL~\cite{gao2023empowered} & 40.19   & 23.93      & 21.12         & 33.57 \\
    & Ours            & 41.26   & 25.05     & 22.53         & 34.48 \\ \hline
    \end{tabular}
}
\end{table}

We conducted experiments to evaluate the few-shot learning performance of open-source LLMs using the ``SFTI-COT-SK-FULL'' approach with the $C_{VDT}$ column definition, specifically focusing on Llama2 and Code Llama (Tables~\ref{tab:llama2_ft_sft_cot} and~\ref{tab:codellama_ft_sft_cot}). Our proposed approach outperformed other example selection methods, showcasing superior performance, and increasing the number of examples resulted in improved performance. However, when comparing these results with those in Figures~\ref{fig:qr_llama2} and \ref{fig:qr_codellama}, a noticeable performance degradation is evident, possibly due to the limited proficiency of both the original Llama2 and Code Llama in Text-to-SQL tasks.

\noindent\textbf{Error Analysis}.

\begin{table}[h]
\centering
\caption{Statistics of failures in Llama2 fine-tuned with the \textbf{SFTI-COT-SK-FULL} method under zero-Shot setting for Text-to-SQL tasks. Error proportions are calculated as the number of errors divided by the total number of queries in BIRD-Dev, and the $C_{VDT}$ column definition is utilized for database schema.
}
\label{tab:app_failures_llama}
\resizebox{0.5\textwidth}{!}{
\begin{tabular}{c|c|c|c}
\hline
                                             & Error categories     & Original & Fine-tuned \\ \hline
\multirow{5}{*}{Wrong schema linking}        & Tables not exist     & 0.65\%  & 0.97\%            \\
                                             & Columns not exist    & 40.4\%   & 4.1\%            \\
                                             & Wrong tables         & 1.3\%  & 3.39\%            \\
                                             & Wrong columns        & 0.98\%    & 8.3\%             \\
                                             & Wrong where statment & 4.9\%    & 23.99\%           \\ \hline
\multirow{2}{*}{Incorrect JOIN operation}    & Wrong tables         & 6.12\%  & 12.84\%           \\
                                             & Wrong columns        & 0.78\%  & 2.4\%            \\ \hline
\multirow{2}{*}{Inaccurate nested structure} & Set operation        & 0.46\%  & 0.65\%            \\
                                             & Wrong sub-query      & 4.4\%  & 2.28\%            \\ \hline
\multirow{2}{*}{Other}                       & Syntax error         & 37.9\%   & 0\%              \\ \cline{2-4} 
                                             & Group-by error       & 0.97\%    & 0.65\%            \\ \hline
\end{tabular}
}
\end{table}

\begin{table}[h]
\centering
\caption{Statistics of failures in Code Llama fine-tuned with the \textbf{SFTI-COT-SK-FULL} method under zero-Shot setting for Text-to-SQL tasks. Error proportions are calculated as the number of errors divided by the total number of queries in BIRD-Dev, and the $C_{VDT}$ column definition is utilized for database schema.}
\label{tab:app_failures_cllama}
\resizebox{0.5\textwidth}{!}{
\begin{tabular}{c|c|c|c}
\hline
                                             & Error categories     & Original & Fine-tuned \\ \hline
\multirow{5}{*}{Wrong schema linking}        & Tables not exist     & 0.71\%  & 0.39\%            \\
                                             & Columns not exist    & 32.14\%   & 3.2\%            \\
                                             & Wrong tables         & 1.11\%  & 2.67\%           \\
                                             & Wrong columns        & 0.78\%  & 6.19\%            \\
                                             & Wrong where statment & 7.8\%    & 21.3\%           \\ \hline
\multirow{2}{*}{Incorrect JOIN operation}    & Wrong tables         & 13.75\% & 10.56\%           \\
                                             & Wrong columns        & 5.4\%  & 2.87\%            \\ \hline
\multirow{2}{*}{Inaccurate nested structure} & Set operation        & 1.37\%  & 0.91\%            \\
                                             & Wrong sub-query      & 6.7\%  & 3.65\%            \\ \hline
\multirow{2}{*}{Other}                       & Syntax error         & 14.6\% & 0\%              \\ \cline{2-4} 
                                             & Group-by error       & 0.84\%    & 0.65\%            \\ \hline
\end{tabular}
}
\end{table}

Tables~\ref{tab:app_failures_llama} and~\ref{tab:app_failures_cllama} exhibit significant reductions in various error categories for both Llama2 and Code Llama following fine-tuning. Particularly, errors such as ``syntax error'' and the majority of ``column not exist'' issues have been substantially mitigated post fine-tuning, suggesting an improved ability to avoid generating SQL statements with obvious syntax errors or referencing non-existing tables and columns. However, challenges persist in selecting appropriate tables and columns, as well as in generating accurate SQL statements for both models.
\subsection{Token Efficiency}
\label{subsec:summary}

Tables~\ref{app:tab_te_zero} and \ref{app:tab_te_few}  display the outcomes of column truncation carried out to conform to the token length limitation of less than 2048, established to meet memory constraints. It is important to note that we input 2048 tokens during the inference stage after fine-tuning for alignment purposes. The results in both tables highlight the significant roles played by our proposed \textbf{Target Column Truncation} and \textbf{Example Column Truncation} in ensuring the success of supervised fine-tuning and few-shot learning for open-source LLMs in Text-to-SQL tasks. Particularly in few-shot learning, the introduction of three examples results in 100\% column truncations for all queries.

\begin{table}[h]
\centering

\caption{Column truncation in supervised fine-tuning. The column labeled ``Average truncated columns'' represents the average number of truncated columns for queries involving column truncation. ``SFT'' indicates supervised fine-tuning without COT, and ``SFT-COT'' indicates supervised fine-tuning with COT.}
\label{app:tab_te_zero}
\resizebox{0.5\textwidth}{!}{
    \begin{tabular}{c|ccc}
    \hline
    & 
    \makecell[l]{Queries with column \\  truncation/Total queries } 
      &
      \makecell[l]{Average truncated\\ columns}
       \\ \hline
    SFT   & 11.68\%                                           & 19.3                                                      \\ \hline
    SFT-COT       & 11.73\%                                           & 20.2                                                      \\ \hline
    \end{tabular}
}
\end{table}

\begin{table}[h]
\centering
\caption{Token efficiency in few-shot learning. The column labeled ``Average truncated target columns'' represents the average number of truncated columns for the target queries in cases involving column truncation, and the column labeled ``Average truncated example columns'' represents the average number of truncated columns for the examples in queries involving column truncation. ``SFT'' indicates supervised fine-tuning without COT, and ``SFT-COT'' indicates supervised fine-tuning with COT.}
\label{app:tab_te_few}
\resizebox{0.5\textwidth}{!}{
    \begin{tabular}{c|cccc}
    \hline
    & No. of shots&
    \makecell[l]{Queries with column \\  truncation/Total queries} 
      &
      \makecell[l]{Average truncated\\ target columns}
       &
      \makecell[l]{Average truncated\\ example columns}
       \\ \hline
    \multirow{2}{*}{
    \makecell[l]{SFT} }
    &0-shot        & 8.42\%                                           & 72.1 &0  \\ 
    &1-shot        & 84.7\%                                           & 24.2 &2.9  \\ 
    &3-shot          & 100\%                                            & 66.7 & 11.0    \\ \hline
    \multirow{2}{*}{
    
    \makecell[l]{SFT-COT} }
    &0-shot        & 8.39\%                                           & 72.3 &0  \\ 
    &1-shot        & 85.4\%                                           & 24.4 &3.2  \\ 
    &3-shot          & 100\%                                            & 66.9 & 11.3    \\ \hline
    \end{tabular}
}
\end{table}
\subsection{Time costs}
\label{subsec:inf_time}

Table~\ref{tab:time} offers a comparative analysis of the time costs associated with six models. Initially, for Code Llama, the inference process required approximately 3.4 seconds. Following supervised fine-tuning, excluding COT, Code Llama's inference time decreased by about 0.6 seconds. This reduction underscores the model's enhanced proficiency in the Text-to-SQL task post-fine-tuning, resulting in a reduction in computational effort and inference time. However, upon further integration of COT, consisting of three distinct sub-tasks, Code Llama's inference time increased to around 6 seconds. Consequently, while COT may enhance performance, it leads to a doubling in inference time.

\begin{table}[h]
\centering
\caption{Comparison of inference time costs of six models, encompassing GPT-3.5-TURBO, original Llama2, original CodeLlama, SFT Code Llama without COT (Chain of Thoughts), and SFT Code Llama with COT (``SFTI-COT-SK-FULL'').}
\label{tab:time}
\begin{tabular}{l|l}
\hline
Model               & Times(s) \\ \hline
GPT-3.5-TURBO              & 1.8s     \\ \hline
Original Llama2     & 3.6s     \\ \hline
Original Code Llama & 3.4s     \\ \hline
SFT Code Llama w/o COT & 2.8s     \\ \hline
SFT Code Llama with COT  & 6.0s       \\ \hline
\end{tabular}
\end{table}

\section{Conclusions}
\label{sec:conclusion}

\begin{table}[]
\centering
\caption{Execution accuracy (EX) on the BIRD dataset, according to the BIRD leaderboard.}
\label{tab:results_com_bird}
\resizebox{0.5\textwidth}{!}{
    \begin{tabular}{|l|l|l|l|}
    \hline
    Rank & Model            & Size&BIRD-Dev (\%)    \\ \hline
    7& \textbf{Open-SQL (Ours)}  & 7B & \textbf{48.24}  \\ \hline
    8 & GPT-4            & UNK & 46.35  \\ \hline
    9 & Claude-2         & UNK & 42.70 \\ \hline
    %10 & \textbf{Open-SQL (Early version)}  & 7B &37.68 & 47.74 \\ \hline
    11 & ChatGPT + CoT & UNK	&	36.64 \\\hline
    12 & ChatGPT          & UNK & 37.72 \\ \hline
    13 & CodeX            & 175B & 34.35 \\ \hline
    14 & Palm-2           & UNK & 27.38 \\ \hline
    18 & T5-3B            & 3B & 23.34 \\ \hline          
    \end{tabular}
}
\end{table}

\begin{table}[!h]
\centering
\caption{Execution accuracy (EX) on the Spider dataset, according to the Spider leaderboard.}
\label{tab:results_com_spider}
\resizebox{0.5\textwidth}{!}{
    \begin{tabular}{|l|l|l|l|}
    \hline
    Rank & Model            & Size&Spider-Dev (\%)    \\ \hline
    1 & DAIL-SQL + GPT-4  & UNK & 86.6  \\ \hline
    4 & DIN-SQL + GPT-4            & UNK & 85.3  \\ \hline
    5 & \textbf{Open-SQL (Ours)}    & 7B & \textbf{84.1} \\ \hline
    6 & Hindsight Chain of Thought with GPT-4 & UNK	&	83.9 \\\hline
    10 & DIN-SQL + CodeX          & UNK & 78.2 \\ \hline
    11 & ChatGPT 3.5            & UNK & 72.3 \\ \hline       
    \end{tabular}
}
\end{table}

This paper systematically evaluates open-source LLMs in Text-to-SQL tasks, introducing the \openprompt strategy for effective question representation and advocating for supervised fine-tuning. Noteworthy contributions include exploring Chain-of-Thought benefits in step-by-step inference, proposing \openexample for enhanced few-shot learning, and introducing token-efficient techniques (\textbf{Variable-length Open DB Schema}, \textbf{Target Column Truncation}, and \textbf{Example Column Truncation}) to address large-scale database challenges. 

Our methodology brings about significant enhancements for both Llama2 and Code Llama, resulting in substantial performance improvements (2.54\% to 41.04\% for Llama2-7B and 14.54\% to 48.24\% for Code Llama-7B). Furthermore, the findings in Table~\ref{tab:results_com_bird} highlight that Code Llama, fine-tuned with a modest 7 billion parameters, surpasses ChatGPT in performance and is on par with GPT-4. Code Llama-7B, fine-tuned using our proposed supervised fine-tuning with Chain-of-Thought (COT) approach, labeled as ``SFTI-COT-SK-FULL'', achieved an impressive \textbf{48.24\%} execution accuracy, surpassing GPT-4's 46.35\%, on the BIRD-Dev dataset. Furthermore, we conducted experiments employing identical procedures on the Spider dataset. As shown in Table ~\ref{tab:results_com_spider}, we achieved an execution accuracy of 84.1\%. While this performance surpassed ChatGPT 3.5's on the same dataset, it fell short of the results obtained with GPT-4. The Spider is characterized by its simplicity, featuring fewer tables and columns within the database. Additionally, the majority of table and column names are explicitly mentioned in the questions, significantly easing the difficulty of linking schema items. In this context, the primary challenge of the Text-to-SQL task revolves around enabling the model to comprehend the input table structure and questions effectively. GPT-4's strong comprehension capability allows it to understand the table structure and questions more effectively than models like Code Llama, resulting in superior performance on the Spider.

\noindent\textbf{Limitations}: Our experimental results underscore the challenges in maintaining the effective learning capabilities of open-source LLMs from contextual examples after fine-tuning, corroborating findings in~\cite{gao2023empowered}. Furthermore, there are difficulties in correctly identifying tables for JOIN operations and generating intricate WHERE statements. These findings underscore the critical necessity for future studies to delve into schema linking and explore methods for preserving in-context learning abilities after fine-tuning.
%This emphasizes the crucial need for further investigations into preserving in-context learning ability after fine-tuning in future studies. 

% \section*{Acknowledgment}

% This research was supported in part by NSFC under Grant no. 92270122; and in part by Guangdong Provincial Natural Science Foundation under grant no. 2023A1515012584; and in part by the Shenzhen Research Foundation for Basic Research, China, under Grant JCYJ20210324093000002

\clearpage
%\section*{References}

\bibliography{custom}

\begin{thebibliography}{33}
\expandafter\ifx\csname natexlab\endcsname\relax\def\natexlab#1{#1}\fi

\bibitem[{Cai et~al.(2018)Cai, Xu, Zhang, Yang, Li, and Liang}]{DBLP:conf/ijcai/CaiXZYLL18}
Ruichu Cai, Boyan Xu, Zhenjie Zhang, Xiaoyan Yang, Zijian Li, and Zhihao Liang. 2018.
\newblock An encoder-decoder framework translating natural language to database queries.
\newblock In \emph{Proceedings of the Twenty-Seventh International Joint Conference on Artificial Intelligence}, pages 3977--3983.

\bibitem[{Chang and Fosler-Lussier(2023)}]{chang2023prompt}
Shuaichen Chang and Eric Fosler-Lussier. 2023.
\newblock \href {http://arxiv.org/abs/2305.11853} {How to prompt llms for text-to-sql: A study in zero-shot, single-domain, and cross-domain settings}.

\bibitem[{Deng et~al.(2022)Deng, Chen, and Zhang}]{deng2022recent}
Naihao Deng, Yulong Chen, and Yue Zhang. 2022.
\newblock Recent advances in text-to-sql: {A} survey of what we have and what we expect.
\newblock In \emph{Proceedings of the 29th International Conference on Computational Linguistics}, pages 2166--2187.

\bibitem[{Dong et~al.(2023{\natexlab{a}})Dong, Li, Dai, Zheng, Wu, Chang, Sun, Xu, Li, and Sui}]{icl22}
Qingxiu Dong, Lei Li, Damai Dai, Ce~Zheng, Zhiyong Wu, Baobao Chang, Xu~Sun, Jingjing Xu, Lei Li, and Zhifang Sui. 2023{\natexlab{a}}.
\newblock A survey for in-context learning.
\newblock \emph{CoRR}, abs/2301.00234.

\bibitem[{Dong et~al.(2023{\natexlab{b}})Dong, Zhang, Ge, Mao, Gao, Chen, Lin, and Lou}]{dong2023c3}
Xuemei Dong, Chao Zhang, Yuhang Ge, Yuren Mao, Yunjun Gao, Lu~Chen, Jinshu Lin, and Dongfang Lou. 2023{\natexlab{b}}.
\newblock {C3:} zero-shot text-to-sql with chatgpt.
\newblock \emph{CoRR}, abs/2307.07306.

\bibitem[{Gao et~al.(2023)Gao, Wang, Li, Sun, Qian, Ding, and Zhou}]{gao2023empowered}
Dawei Gao, Haibin Wang, Yaliang Li, Xiuyu Sun, Yichen Qian, Bolin Ding, and Jingren Zhou. 2023.
\newblock Text-to-sql empowered by large language models: A benchmark evaluation.
\newblock \emph{CoRR}, abs/2308.15363.

\bibitem[{Guo et~al.(2023)Guo, Tian, Tang, Wang, Wen, Yang, and Wang}]{guo2023framework}
Chunxi Guo, Zhiliang Tian, Jintao Tang, Pancheng Wang, Zhihua Wen, Kang Yang, and Ting Wang. 2023.
\newblock A case-based reasoning framework for adaptive prompting in cross-domain text-to-sql.
\newblock \emph{CoRR}, abs/2304.13301.

\bibitem[{Guo et~al.(2019)Guo, Zhan, Gao, Xiao, Lou, Liu, and Zhang}]{DBLP:conf/acl/GuoZGXLLZ19}
Jiaqi Guo, Zecheng Zhan, Yan Gao, Yan Xiao, Jian{-}Guang Lou, Ting Liu, and Dongmei Zhang. 2019.
\newblock Towards complex text-to-sql in cross-domain database with intermediate representation.
\newblock In \emph{Proceedings of the 57th Conference of the Association for Computational Linguistics}, pages 4524--4535.

\bibitem[{Hu et~al.(2021)Hu, Shen, Wallis, Allen-Zhu, Li, Wang, Wang, and Chen}]{hu2021lora}
Edward~J. Hu, Yelong Shen, Phillip Wallis, Zeyuan Allen-Zhu, Yuanzhi Li, Shean Wang, Lu~Wang, and Weizhu Chen. 2021.
\newblock \href {http://arxiv.org/abs/2106.09685} {Lora: Low-rank adaptation of large language models}.

\bibitem[{Hui et~al.(2022)Hui, Geng, Wang, Qin, Li, Li, Sun, and Li}]{DBLP:conf/acl/HuiGWQLLSL22}
Binyuan Hui, Ruiying Geng, Lihan Wang, Bowen Qin, Yanyang Li, Bowen Li, Jian Sun, and Yongbin Li. 2022.
\newblock S{\({^2}\)}sql: Injecting syntax to question-schema interaction graph encoder for text-to-sql parsers.
\newblock In \emph{Findings of the Association for Computational Linguistics}, pages 1254--1262.

\bibitem[{Lei et~al.(2020)Lei, Wang, Ma, Gan, Lu, Kan, and Chua}]{lei2020re}
Wenqiang Lei, Weixin Wang, Zhixin Ma, Tian Gan, Wei Lu, Min-Yen Kan, and Tat-Seng Chua. 2020.
\newblock Re-examining the role of schema linking in text-to-sql.
\newblock In \emph{Proceedings of the 2020 Conference on Empirical Methods in Natural Language Processing (EMNLP)}, pages 6943--6954.

\bibitem[{Li et~al.(2023{\natexlab{a}})Li, Zhang, Li, and Chen}]{li2023resdsql}
Haoyang Li, Jing Zhang, Cuiping Li, and Hong Chen. 2023{\natexlab{a}}.
\newblock Resdsql: Decoupling schema linking and skeleton parsing for text-to-sql.
\newblock AAAI-23, page 13067–13075. AAAI Press.

\bibitem[{Li et~al.(2023{\natexlab{b}})Li, Hui, Cheng, Qin, Ma, Huo, Huang, Du, Si, and Li}]{li2023graphix}
Jinyang Li, Binyuan Hui, Reynold Cheng, Bowen Qin, Chenhao Ma, Nan Huo, Fei Huang, Wenyu Du, Luo Si, and Yongbin Li. 2023{\natexlab{b}}.
\newblock Graphix-t5: Mixing pre-trained transformers with graph-aware layers for text-to-sql parsing.
\newblock In \emph{37th {AAAI} Conference on Artificial Intelligence}, pages 13076--13084.

\bibitem[{Li et~al.(2023{\natexlab{c}})Li, Hui, Qu, Li, Yang, Li, Wang, Qin, Cao, Geng et~al.}]{DBLP:journals/corr/abs-2305-03111}
Jinyang Li, Binyuan Hui, Ge~Qu, Binhua Li, Jiaxi Yang, Bowen Li, Bailin Wang, Bowen Qin, Rongyu Cao, Ruiying Geng, et~al. 2023{\natexlab{c}}.
\newblock Can {LLM} already serve as {A} database interface? {A} big bench for large-scale database grounded text-to-sqls.
\newblock \emph{CoRR}, abs/2305.03111.

\bibitem[{Liu et~al.(2023{\natexlab{a}})Liu, Hu, Wen, and Yu}]{DBLP:journals/corr/abs-2303-13547}
Aiwei Liu, Xuming Hu, Lijie Wen, and Philip~S. Yu. 2023{\natexlab{a}}.
\newblock A comprehensive evaluation of chatgpt's zero-shot text-to-sql capability.
\newblock \emph{CoRR}, abs/2303.13547.

\bibitem[{Liu et~al.(2023{\natexlab{b}})Liu, Shi, Zhang, Wang, Li, and Kong}]{DBLP:conf/ijcnn/LiuSZWLK23}
Hu~Liu, Yuliang Shi, Jianlin Zhang, Xinjun Wang, Hui Li, and Fanyu Kong. 2023{\natexlab{b}}.
\newblock Multi-hop relational graph attention network for text-to-sql parsing.
\newblock In \emph{International Joint Conference on Neural Networks}, pages 1--8.

\bibitem[{Liu et~al.(2022)Liu, Shen, Zhang, Dolan, Carin, and Chen}]{liu2022in-context}
Jiachang Liu, Dinghan Shen, Yizhe Zhang, Bill Dolan, Lawrence Carin, and Weizhu Chen. 2022.
\newblock What makes good in-context examples for gpt-3?
\newblock In \emph{Proceedings of Deep Learning Inside Out: The 3rd Workshop on Knowledge Extraction and Integration for Deep Learning Architectures}, pages 100--114.

\bibitem[{Nan et~al.(2023)Nan, Zhao, Zou, Ri, Tae, Zhang, Cohan, and Radev}]{nan2023enhancing}
Linyong Nan, Yilun Zhao, Weijin Zou, Narutatsu Ri, Jaesung Tae, Ellen Zhang, Arman Cohan, and Dragomir Radev. 2023.
\newblock Enhancing few-shot text-to-sql capabilities of large language models: {A} study on prompt design strategies.
\newblock \emph{CoRR}, abs/2305.12586.

\bibitem[{OpenAI(2023{\natexlab{a}})}]{gpt4}
OpenAI. 2023{\natexlab{a}}.
\newblock {GPT-4} technical report.
\newblock \emph{CoRR}, abs/2303.08774.

\bibitem[{OpenAI(2023{\natexlab{b}})}]{chatgpt}
OpenAI. 2023{\natexlab{b}}.
\newblock Introducing chatgpt.
\newblock \url{https://openai.com/blog/chatgpt}.
\newblock Last accessed on 2023-07-24.

\bibitem[{OpenAI(2023{\natexlab{c}})}]{openaiprompt}
OpenAI. 2023{\natexlab{c}}.
\newblock Sql translate.
\newblock \url{https://platform.openai.com/examples/default-sql-translate}.
\newblock Last accessed on 2023-07-24.

\bibitem[{Popescu et~al.(2022)Popescu, Manotas, Vo, Yeo, Khorashani, and Sheinin}]{DBLP:conf/coling/PopescuMVYKS22}
Octavian Popescu, Irene Manotas, Ngoc Phuoc~An Vo, Hangu Yeo, Elahe Khorashani, and Vadim Sheinin. 2022.
\newblock Addressing limitations of encoder-decoder based approach to text-to-sql.
\newblock In \emph{Proceedings of the 29th International Conference on Computational Linguistics}, pages 1593--1603.

\bibitem[{Pourreza and Rafiei(2023)}]{pourreza2023dinsql}
Mohammadreza Pourreza and Davood Rafiei. 2023.
\newblock \href {http://arxiv.org/abs/2304.11015} {Din-sql: Decomposed in-context learning of text-to-sql with self-correction}.

\bibitem[{Qi et~al.(2022)Qi, Tang, He, Wan, Cheng, Zhou, Wang, Zhang, and Lin}]{qi2022rasat}
Jiexing Qi, Jingyao Tang, Ziwei He, Xiangpeng Wan, Yu~Cheng, Chenghu Zhou, Xinbing Wang, Quanshi Zhang, and Zhouhan Lin. 2022.
\newblock {RASAT:} integrating relational structures into pretrained seq2seq model for text-to-sql.
\newblock In \emph{Proceedings of the 2022 Conference on Empirical Methods in Natural Language Processing}, pages 3215--3229.

\bibitem[{Scholak et~al.(2021)Scholak, Schucher, and Bahdanau}]{scholak2021picard}
Torsten Scholak, Nathan Schucher, and Dzmitry Bahdanau. 2021.
\newblock {PICARD:} parsing incrementally for constrained auto-regressive decoding from language models.
\newblock In \emph{Proceedings of the 2021 Conference on Empirical Methods in Natural Language Processing}, pages 9895--9901.

\bibitem[{Sun et~al.(2023)Sun, Arik, Nakhost, Dai, Sinha, Yin, and Pfister}]{sun2023sql}
Ruoxi Sun, Sercan~{\"{O}}. Arik, Hootan Nakhost, Hanjun Dai, Rajarishi Sinha, Pengcheng Yin, and Tomas Pfister. 2023.
\newblock Sql-palm: Improved large language model adaptation for text-to-sql.
\newblock \emph{CoRR}, abs/2306.00739.

\bibitem[{Touvron et~al.(2023{\natexlab{a}})Touvron, Lavril, Izacard, Martinet, Lachaux, Lacroix, Rozi{\`{e}}re, Goyal, Hambro, Azhar, Rodriguez, Joulin, Grave, and Lample}]{llama}
Hugo Touvron, Thibaut Lavril, Gautier Izacard, Xavier Martinet, Marie{-}Anne Lachaux, Timoth{\'{e}}e Lacroix, Baptiste Rozi{\`{e}}re, Naman Goyal, Eric Hambro, Faisal Azhar, Aur{\'{e}}lien Rodriguez, Armand Joulin, Edouard Grave, and Guillaume Lample. 2023{\natexlab{a}}.
\newblock Llama: Open and efficient foundation language models.
\newblock \emph{CoRR}, abs/2302.13971.

\bibitem[{Touvron et~al.(2023{\natexlab{b}})Touvron, Martin, Stone, Albert, Almahairi, Babaei, Bashlykov, Batra, Bhargava, Bhosale, Bikel, Blecher, Ferrer, Chen, Cucurull, Esiobu, Fernandes, Fu, Fu, Fuller, Gao, Goswami, Goyal, Hartshorn, Hosseini, Hou, Inan, Kardas, Kerkez, Khabsa, Kloumann, Korenev, Koura, Lachaux, Lavril, Lee, Liskovich, Lu, Mao, Martinet, Mihaylov, Mishra, Molybog, Nie, Poulton, Reizenstein, Rungta, Saladi, Schelten, Silva, Michael, Ranjan, Xiaoqing, Tan, Tang, Taylor, Williams, Kuan, Xu, Yan, Zarov, Zhang, Fan, Kambadur, Narang, Rodriguez, Stojnic, Edunov, and Scialom}]{llama2}
Hugo Touvron, Louis Martin, Kevin Stone, Peter Albert, Amjad Almahairi, Yasmine Babaei, Nikolay Bashlykov, Soumya Batra, Prajjwal Bhargava, Shruti Bhosale, Dan Bikel, Lukas Blecher, Cristian~Canton Ferrer, Moya Chen, Guillem Cucurull, David Esiobu, Jude Fernandes, Jeremy Fu, Wenyin Fu, Brian Fuller, Cynthia Gao, Vedanuj Goswami, Naman Goyal, Anthony Hartshorn, Saghar Hosseini, Rui Hou, Hakan Inan, Marcin Kardas, Viktor Kerkez, Madian Khabsa, Isabel Kloumann, Artem Korenev, Singh Koura, Marie-Anne Lachaux, Thibaut Lavril, Jenya Lee, Diana Liskovich, Yinghai Lu, Yuning Mao, Xavier Martinet, Todor Mihaylov, Pushkar Mishra, Igor Molybog, Yixin Nie, Andrew Poulton, Jeremy Reizenstein, Rashi Rungta, Kalyan Saladi, Alan Schelten, Ruan Silva, Eric Michael, Smith Ranjan, Subramanian Xiaoqing, Ellen Tan, Binh Tang, Ross Taylor, Adina Williams, Jian~Xiang Kuan, Puxin Xu, Zheng Yan, Iliyan Zarov, Yuchen Zhang, Angela Fan, Melanie Kambadur, Sharan Narang, Aurelien Rodriguez, Robert Stojnic, Sergey Edunov, and Thomas
  Scialom. 2023{\natexlab{b}}.
\newblock Llama2: Open foundation and fine-tuned chat models.
\newblock \emph{CoRR}.

\bibitem[{Wang et~al.(2020)Wang, Shin, Liu, Polozov, and Richardson}]{rat-sql}
Bailin Wang, Richard Shin, Xiaodong Liu, Oleksandr Polozov, and Matthew Richardson. 2020.
\newblock {RAT-SQL:} relation-aware schema encoding and linking for text-to-sql parsers.
\newblock In \emph{Proceedings of the 58th Annual Meeting of the Association for Computational Linguistics}, pages 7567--7578.

\bibitem[{Wang et~al.(2022)Wang, Qin, Hui, Li, Yang, Wang, Li, Sun, Huang, Si, and Li}]{wang2022proton}
Lihan Wang, Bowen Qin, Binyuan Hui, Bowen Li, Min Yang, Bailin Wang, Binhua Li, Jian Sun, Fei Huang, Luo Si, and Yongbin Li. 2022.
\newblock Proton: Probing schema linking information from pre-trained language models for text-to-sql parsing.
\newblock In \emph{The 28th {ACM} {SIGKDD} Conference on Knowledge Discovery and Data Mining}, pages 1889--1898.

\bibitem[{Xu et~al.(2018)Xu, Wu, Wang, Feng, and Sheinin}]{DBLP:conf/emnlp/XuWWFS18}
Kun Xu, Lingfei Wu, Zhiguo Wang, Yansong Feng, and Vadim Sheinin. 2018.
\newblock Sql-to-text generation with graph-to-sequence model.
\newblock In \emph{Proceedings of the 2018 Conference on Empirical Methods in Natural Language Processing}, pages 931--936.

\bibitem[{Yin et~al.(2020)Yin, Neubig, Yih, and Riedel}]{yin2020tabert}
Pengcheng Yin, Graham Neubig, Wen-tau Yih, and Sebastian Riedel. 2020.
\newblock Tabert: Pretraining for joint understanding of textual and tabular data.
\newblock \emph{Cornell University - arXiv,Cornell University - arXiv}.

\bibitem[{Zheng et~al.(2022)Zheng, Wang, Dong, Wang, and Li}]{DBLP:conf/acl/ZhengWDWL22}
Yanzhao Zheng, Haibin Wang, Baohua Dong, Xingjun Wang, and Changshan Li. 2022.
\newblock {HIE-SQL:} history information enhanced network for context-dependent text-to-sql semantic parsing.
\newblock In \emph{Findings of the Association for Computational Linguistics}, pages 2997--3007.

\end{thebibliography}

\clearpage
\onecolumn
\appendix

\section{Prompt for Chain-of-Thought}
\label{app:promt_cot}

\begin{lstlisting}[language=Schema, caption={The simple COT template (\textbf{COT\_SP}). \textbf{Tables\_related} encompasses tables pertinent to the task, while \textbf{Columns\_related} includes related columns (paired with their respective tables) likely to be employed in the specified task.}, label={lst:col_simple}]
1.
### SQLite SQL tables are requested to be represented in the following format.
# TABLE_NAME (
# COLUMN_NAME: DESCRIPTION
# )
# FOREIGN KEYS:
# TABLE_NAME1.COLUMN_NAME1 = TABLE_NAME2.COLUMN_NAME2
### Here are SQLite SQL tables, with their properties:
# ${DATABASE SCHEMA}
### Question: ${TARGET_QUESTION}.
### Note that: ${EXTERNAL_KNOWLEDGE}.
Please generate the SQL script STEP BY STEP.
Find the required tables based on the QUESTION.

Tables: (table
         table)

2. 
### SQLite SQL tables are requested to be represented in the following format.
# TABLE_NAME (
# COLUMN_NAME: TYPE, (DESCRIPTION), (VALUE1, VALUE2, ...) 
# )
# FOREIGN KEYS:
# TABLE_NAME1.COLUMN_NAME1 = TABLE_NAME2.COLUMN_NAME2
### Here are SQLite SQL tables that will be used, with their properties:
# ${DATABASE SCHEMA}
### Question: ${TARGET_QUESTION}.
### Note that: ${EXTERNAL_KNOWLEDGE}.
Please generate the SQL script STEP BY STEP.
Given the tables: 
${Tables_related}.
From the given tables, find the required columns based on the QUESTION.

Columns: table: (column
                column)
                
3. 
### SQLite SQL tables are requested to be represented in the following format.
# TABLE_NAME (
# COLUMN_NAME: TYPE, (DESCRIPTION), (VALUE1, VALUE2, ...) 
# )
# FOREIGN KEYS:
# TABLE_NAME1.COLUMN_NAME1 = TABLE_NAME2.COLUMN_NAME2
### Here are SQLite SQL tables that will be used, with their properties:
# ${DATABASE SCHEMA}
### Question: ${TARGET_QUESTION}.
### Note that: ${EXTERNAL_KNOWLEDGE}.
Please generate the SQL script STEP BY STEP.
Given the tables and columns used in the SQL query: 
${Columns_related}.
### Complete sqlite SQL query based on the given tables and columns
SELECT
\end{lstlisting}

\begin{lstlisting}[language=Schema, caption={The skeleton-based Chain of Thought template (\textbf{COT\_SK}) aids in schema linking and SQL generation. The \textbf{Tables\_related} section comprises tables relevant to the task, while \textbf{Columns\_related} includes associated columns (paired with their respective tables) likely to be utilized in the specified task. The resulting \textbf{Skeleton} represents the generated framework.}, label={lst:col_skeleton}]
1.
### SQLite SQL tables are requested to be represented in the following format.
# TABLE_NAME (
# COLUMN_NAME: DESCRIPTION   
# )
# FOREIGN KEYS:
# TABLE_NAME1.COLUMN_NAME1 = TABLE_NAME2.COLUMN_NAME2
### Here are SQLite SQL tables, with their properties:
# ${DATABASE SCHEMA}
### Question: ${TARGET_QUESTION}.
### Note that: ${EXTERNAL_KNOWLEDGE}.
Please generate the SQL script STEP BY STEP.
Find the required tables based on the QUESTION.

Tables: (table
         table)

2. 
### SQLite SQL tables are requested to be represented in the following format.
# TABLE_NAME (
# COLUMN_NAME: TYPE, (DESCRIPTION), (VALUE1, VALUE2, ...) 
# )
# FOREIGN KEYS:
# TABLE_NAME1.COLUMN_NAME1 = TABLE_NAME2.COLUMN_NAME2
### Here are SQLite SQL tables that will be used, with their properties:
# ${DATABASE SCHEMA}
### Question: ${TARGET_QUESTION}.
### Note that: ${EXTERNAL_KNOWLEDGE}.
Please generate the SQL script STEP BY STEP.
Given the tables: 
${Tables_related}.
From the given tables, find the required columns based on the QUESTION.

Columns: table: (column
                column)

                
3.
### SQLite SQL tables are requested to be represented in the following format.
# TABLE_NAME (
# COLUMN_NAME: TYPE, (DESCRIPTION), (VALUE1, VALUE2, ...) 
# )
# FOREIGN KEYS:
# TABLE_NAME1.COLUMN_NAME1 = TABLE_NAME2.COLUMN_NAME2
### Here are SQLite SQL tables that will be used, with their properties:
# ${DATABASE SCHEMA}
### Question: ${TARGET_QUESTION}.
### Note that: ${EXTERNAL_KNOWLEDGE}.
Please generate the SQL script STEP BY STEP.
Given the tables and columns: 
${Columns_related}.
Based on the given the tables and columns, write the skeleton of the SQL query corresponding to the question.


4. 
### SQLite SQL tables are requested to be represented in the following format.
# TABLE_NAME (
# COLUMN_NAME: TYPE, (DESCRIPTION), (VALUE1, VALUE2, ...) 
# )
# FOREIGN KEYS:
# TABLE_NAME1.COLUMN_NAME1 = TABLE_NAME2.COLUMN_NAME2
### Here are SQLite SQL tables that will be used, with their properties:
# ${DATABASE SCHEMA}
### Question: ${TARGET_QUESTION}.
### Note that: ${EXTERNAL_KNOWLEDGE}.
Please generate the SQL script STEP BY STEP.
Given the tables and columns: 
${Columns_related},
and sql skeleton:
${Skeleton}.
### Complete sqlite SQL query based on the given tables, columns and sql_skeleton
SELECT
\end{lstlisting}

\section{Prompt for few-shot learning}
\label{app:promt_shot}

\begin{lstlisting}[language=Schema, label={lst:org_os}, caption={Prompt for few-show learning.}, label={lst:prompt_few_shot}]
### Complete sqlite SQL query only and with no explanation
### SQLite SQL tables are requested to be represented in the following format.
# TABLE_NAME (
# COLUMN_NAME: (DESCRIPTION), (ENUM_VALUE, ENUM_VALUE2, ...), PRIMARY_KEY
# COLUMN_NAME: (DESCRIPTION), (ENUM_VALUE, ENUM_VALUE2, ...)   
# )
# FOREIGN KEYS:
# TABLE_NAME1.COLUMN_NAME1 = TABLE_NAME2.COLUMN_NAME2

### Here are SQLite SQL tables, with their properties:
# ${EXAMPLE DATABASE SCHEMA}
### Question: ${EXAMPLE_QUESTION}.
### Note that: ${EXTERNAL_KNOWLEDGE}.
### Using valid SQLite, answer the following questions for the tables provided above.
${EXAMPLE_SQL}

### Here are SQLite SQL tables, with their properties:
# ${EXAMPLE DATABASE SCHEMA}
### Question: ${EXAMPLE_QUESTION}.
### Note that: ${EXTERNAL_KNOWLEDGE}.
### Using valid SQLite, answer the following questions for the tables provided above.
${EXAMPLE_SQL}

### Here are SQLite SQL tables, with their properties:
# ${EXAMPLE DATABASE SCHEMA}
### Question: ${EXAMPLE_QUESTION}.
### Note that: ${EXTERNAL_KNOWLEDGE}.
### Using valid SQLite, answer the following questions for the tables provided above.
${EXAMPLE_SQL}

### Here are SQLite SQL tables, with their properties:
# ${TARGET_DATABASE_SCHEMA}
### Using valid SQLite , answer the following questions for the tables provided above.
### Question: ${TARGET_QUESTION}.
### Note that: ${EXTERNAL_KNOWLEDGE}.
SELECT
\end{lstlisting}

\section{Prompts for different column definitions}
\label{app:prompt:columns}

\begin{lstlisting}[language=Schema, caption={Prompt for $C_{N}$ with none of the optional column elements.}, label={lst:prompt_cn}]
### Complete sqlite SQL query only and with no explanation
### SQLite SQL tables are requested to be represented in the following format.
# TABLE_NAME (
# COLUMN_NAME   
# )
# FOREIGN KEYS:
# TABLE_NAME1.COLUMN_NAME1 = TABLE_NAME2.COLUMN_NAME2
### Here are SQLite SQL tables, with their properties:
# ${DATABASE SCHEMA}
### Question: ${TARGET_QUESTION}.
### Note that: ${EXTERNAL_KNOWLEDGE}.
SELECT
\end{lstlisting}

\begin{lstlisting}[language=Schema, caption={Prompt for $C_{T}$ with only \textbf{TYPE}.}, label={lst:prompt_ct}]
### Complete sqlite SQL query only and with no explanation
### SQLite SQL tables are requested to be represented in the following format.
# TABLE_NAME (
# COLUMN_NAME: TYPE   
# )
# FOREIGN KEYS:
# TABLE_NAME1.COLUMN_NAME1 = TABLE_NAME2.COLUMN_NAME2
### Here are SQLite SQL tables, with their properties:
# ${DATABASE SCHEMA}
### Question: ${TARGET_QUESTION}.
### Note that: ${EXTERNAL_KNOWLEDGE}.
SELECT
\end{lstlisting}

\begin{lstlisting}[language=Schema, caption={Prompt for $C_{D}$ with column \textbf{DESCRIPTION}.}, label={lst:prompt_cd}]
### Complete sqlite SQL query only and with no explanation
### SQLite SQL tables are requested to be represented in the following format.
# TABLE_NAME (
# COLUMN_NAME: DESCRIPTION   
# )
# FOREIGN KEYS:
# TABLE_NAME1.COLUMN_NAME1 = TABLE_NAME2.COLUMN_NAME2
### Here are SQLite SQL tables, with their properties:
# ${DATABASE SCHEMA}
### Question: ${TARGET_QUESTION}.
### Note that: ${EXTERNAL_KNOWLEDGE}.
SELECT
\end{lstlisting}

\begin{lstlisting}[language=Schema, caption={Prompt for $C_{V}$ with column \textbf{VALUES}.}, label={lst:prompt_cv}]
### Complete sqlite SQL query only and with no explanation
### SQLite SQL tables are requested to be represented in the following format.
# TABLE_NAME (
# COLUMN_NAME: (ENUM_VALUE, ENUM_VALUE2, ...)  
# )
# FOREIGN KEYS:
# TABLE_NAME1.COLUMN_NAME1 = TABLE_NAME2.COLUMN_NAME2
### Here are SQLite SQL tables, with their properties:
# ${DATABASE SCHEMA}
### Question: ${TARGET_QUESTION}.
### Note that: ${EXTERNAL_KNOWLEDGE}.
SELECT
\end{lstlisting}

\begin{lstlisting}[language=Schema, caption={Prompt for $C_{P}$ with column \textbf{PRIMARY\_KEY}.}, label={lst:prompt_cp}]
### Complete sqlite SQL query only and with no explanation
### SQLite SQL tables are requested to be represented in the following format.
# TABLE_NAME (
# COLUMN_NAME: PRIMARY_KEY
# COLUMN_NAME:
# )
# FOREIGN KEYS:
# TABLE_NAME1.COLUMN_NAME1 = TABLE_NAME2.COLUMN_NAME2
### Here are SQLite SQL tables, with their properties:
# ${DATABASE SCHEMA}
### Question: ${TARGET_QUESTION}.
### Note that: ${EXTERNAL_KNOWLEDGE}.
SELECT
\end{lstlisting}

\begin{lstlisting}[language=Schema, caption={Prompt for $C_{VD}$ with column \textbf{DESCRIPTION} and \textbf{VALUES}.}, label={lst:prompt_cvd}]
### Complete sqlite SQL query only and with no explanation
### SQLite SQL tables are requested to be represented in the following format.
# TABLE_NAME (
# COLUMN_NAME: (DESCRIPTION), (ENUM_VALUE, ENUM_VALUE2, ...)  
# )
# FOREIGN KEYS:
# TABLE_NAME1.COLUMN_NAME1 = TABLE_NAME2.COLUMN_NAME2
### Here are SQLite SQL tables, with their properties:
# ${DATABASE SCHEMA}
### Question: ${TARGET_QUESTION}.
### Note that: ${EXTERNAL_KNOWLEDGE}.
SELECT
\end{lstlisting}

\begin{lstlisting}[language=Schema, caption={Prompt for $C_{VDT}$ with column \textbf{DESCRIPTION}, \textbf{VALUES}, and \textbf{TYPE}.}, label={lst:prompt_cvdt}]
### Complete sqlite SQL query only and with no explanation
### SQLite SQL tables are requested to be represented in the following format.
# TABLE_NAME (
# COLUMN_NAME: TYPE, (DESCRIPTION), (ENUM_VALUE, ENUM_VALUE2, ...)   
# )
# FOREIGN KEYS:
# TABLE_NAME1.COLUMN_NAME1 = TABLE_NAME2.COLUMN_NAME2
### Here are SQLite SQL tables, with their properties:
# ${DATABASE SCHEMA}
### Question: ${TARGET_QUESTION}.
### Note that: ${EXTERNAL_KNOWLEDGE}.
SELECT
\end{lstlisting}

\begin{lstlisting}[language=Schema, caption={Prompt for $C_{VDT}$ with column \textbf{DESCRIPTION}, \textbf{VALUES}, and \textbf{PRIMARY\_KEY}.}, label={lst:prompt_cvdp}]
### Complete sqlite SQL query only and with no explanation
### SQLite SQL tables are requested to be represented in the following format.
# TABLE_NAME (
# COLUMN_NAME: (DESCRIPTION), (ENUM_VALUE, ENUM_VALUE2, ...), PRIMARY_KEY
# COLUMN_NAME: (DESCRIPTION), (ENUM_VALUE, ENUM_VALUE2, ...)   
# )
# FOREIGN KEYS:
# TABLE_NAME1.COLUMN_NAME1 = TABLE_NAME2.COLUMN_NAME2
### Here are SQLite SQL tables, with their properties:
# ${DATABASE SCHEMA}
### Question: ${TARGET_QUESTION}.
### Note that: ${EXTERNAL_KNOWLEDGE}.
SELECT
\end{lstlisting}

\begin{lstlisting}[language=Schema, caption={Prompt for $C_{A}$ with column \textbf{TYPE}, \textbf{DESCRIPTION}, \textbf{VALUES}, and \textbf{PRIMARY\_KEY}.}, label={lst:prompt_ca}]
### Complete sqlite SQL query only and with no explanation
### SQLite SQL tables are requested to be represented in the following format.
# TABLE_NAME (
# COLUMN_NAME: TYPE, (DESCRIPTION), (ENUM_VALUE, ENUM_VALUE2, ...), PRIMARY_KEY
# COLUMN_NAME: TYPE, (DESCRIPTION), (ENUM_VALUE, ENUM_VALUE2, ...)    
# )
# FOREIGN KEYS:
# TABLE_NAME1.COLUMN_NAME1 = TABLE_NAME2.COLUMN_NAME2
### Here are SQLite SQL tables, with their properties:
# ${DATABASE SCHEMA}
### Question: ${TARGET_QUESTION}.
### Note that: ${EXTERNAL_KNOWLEDGE}.
SELECT
\end{lstlisting}

\end{document}